%% file: main.tex
\documentclass[letterpaper, 10 pt, conference]{ieeeconf}
\IEEEoverridecommandlockouts
\usepackage{amsmath,amssymb,amsfonts}
\usepackage{algorithmic}
\usepackage{graphicx}
\usepackage{textcomp}
\usepackage[x11names,table]{xcolor}
\usepackage{booktabs}
\usepackage{multirow}
\usepackage{subcaption}
\usepackage{xspace}
\usepackage{paralist}
\makeatletter
\let\NAT@parse\undefined
\makeatother
\usepackage{hyperref}
\hypersetup{colorlinks=true, linkcolor=red, citecolor=blue, urlcolor=magenta}
\usepackage{tablefootnote}

\usepackage[table]{xcolor}

\newcommand{\green}[1]{\textcolor[RGB]{96,177,87}{#1}}
\newcommand{\fn}[1]{{#1}}
\newcommand{\gbf}[1]{\green{\bf{\fn{(#1)}}}}
\newcommand{\rbf}[1]{\textcolor{gray}{\bf{\fn{#1}}}}
\definecolor{graytablerow}{gray}{0.5}

\usepackage{pifont}

\definecolor{darkpastelred}{rgb}{0.76, 0.23, 0.13}
\definecolor{darkcyan}{rgb}{0.0, 0.55, 0.55}
\definecolor{darkmagenta}{rgb}{0.55, 0.0, 0.55}

\definecolor{RoyalBlue}{RGB}{65, 105, 225}
\definecolor{PineGreen}{RGB}{1, 121, 111}
\definecolor{YellowOrange}{RGB}{255, 174, 66}

\makeatletter
\DeclareRobustCommand\onedot{\futurelet\@let@token\@onedot}
\def\@onedot{\ifx\@let@token.\else.\null\fi\xspace}
\def\eg{\emph{e.g}\onedot} 
\def\ie{\emph{i.e}\onedot}

\def\etal{\emph{et al}\onedot}
\makeatother

\begin{document}
\bstctlcite{IEEEexample:BSTcontrol} 

\title{\LARGE \bf
Fourier Prompt Tuning for Modality-Incomplete Scene Segmentation
\author{Ruiping Liu$^{1}$, Jiaming Zhang$^{1,\dag}$, Kunyu Peng$^{1}$, Yufan Chen$^{1}$, Ke Cao$^{1}$, Junwei Zheng$^{1}$,\\M. Saquib Sarfraz$^{1,2}$, Kailun Yang$^{3,4}$, and Rainer Stiefelhagen$^{1}$
}
\thanks{$^{1}$Institute for Anthropomatics and Robotics, Karlsruhe Institute of Technology, Germany.}
\thanks{$^{2}$Mercedes-Benz Tech Innovation, Germany}
\thanks{$^{3}$School of Robotics, Hunan University, China.}
\thanks{$^{4}$National Engineering Research Center of Robot Visual Perception and Control Technology, Hunan University, China.}
\thanks{$^{\dag}$Correspondence: jiaming.zhang@kit.edu}
}
\maketitle

\begin{abstract}
\input{tex/abstract}

\end{abstract}

\section{Introduction}
\input{tex/introduction}

\input{tex/contributions}

\section{Related Work}
\input{tex/related}

\section{Methodology}
\label{methodology}

In this paper, we focus on the parameter-efficient adaptation of multi-modal models to downstream tasks while achieving robustness against Modality-Incomplete Scene Segmentation (MISS). For this purpose, we introduce two key methods: a novel training strategy termed Missing-aware Modal Switch (MMS) for effective data augmentation, and a prompt tuning approach denoted as Fourier Prompt Tuning (FPT), aimed at employing a uniform set of prompts to address diverse missing conditions.

\subsection{Missing-aware Modal Switch}
In semantic segmentation, a subset of dense prediction problems, we handle dense and sparse modalities differently. Effective performance in semantic segmentation requires at least one undamaged dense modality.
Considering a real-world scenario with $n$ dense modalities (\eg, RGB and Depth) and $m$ sparse modalities (\eg, LiDAR and Event), the number of possible modality combinations $C$ can be calculated using Eq.~(\ref{eq:combination}):
\begin{equation}\label{eq:combination}
    C=\sum_{i=1}^n\binom{n}{i}\times\sum_{j=0}^m\binom{m}{j}.
\end{equation}
Associating each data sample with $C$ conditions poses a significant challenge. 
Lee~\etal~\cite{lee2023missing_prompt} address this issue by manually predefining a missing ratio for the entire dataset and identifying the corresponding missing modalities for each sample before training.
In this case, the missing ratio acts as a crucial hyperparameter requiring adjustment and may lead to a more pronounced reliance on specific modalities when improperly defined. Furthermore, the number of samples remains constant (denoted as $d$ samples for the training set) but with different missing situations.
\begin{figure}
    \centering
    \includegraphics[width=1.0\columnwidth]
    {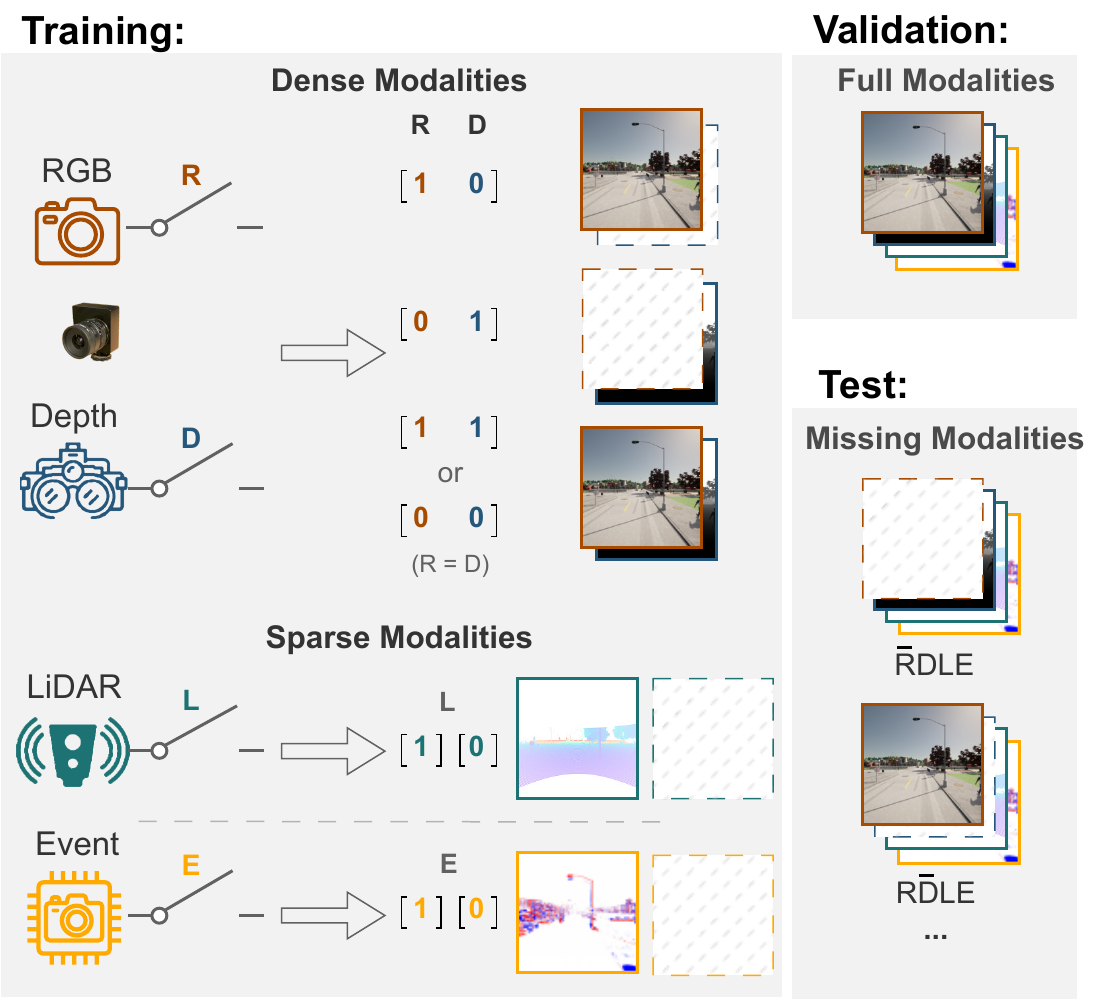}
    \vskip -1ex
    \caption{\small \textbf{Missing-aware Modal Switch (MMS)} method to manage the absence of dense (\eg, \textcolor{brown}{RGB} and \textcolor{RoyalBlue}{Depth}) or/and sparse (\eg, \textcolor{PineGreen}{LiDAR} and \textcolor{YellowOrange}{Event}) modalities. Due to dense prediction, at least one dense modality is retained during training, while modalities are complete during validation and incomplete during testing. The overline on a modality, \eg $\overline{\mathrm{R}}$, means that it is missing.
    }
    \label{fig:MMS}
    \vskip -2ex
\end{figure}
To mitigate the predominant modality reliance resulting from manually defining the missing ratio, we have devised a straightforward training strategy, \textit{Missing-aware Modal Switch (MMS)}, for handling missing modalities, as depicted in Fig.~\ref{fig:MMS}. 
This strategy entails the use of randomly assigned binary switches to govern the presence or absence of each modality. 
A value of `1' denotes that the switch is in the `on' position, while `0' signifies that the switch is in the `off' position.

These binary switches are employed independently for dense and sparse modalities. 
In tasks requiring dense prediction, such as semantic segmentation, the presence of at least one dense modality is crucial. This is because the goal is to predict the category of each pixel.
For this purpose, if all $m$-bits corresponding to dense modalities are set to `0', they are automatically interpreted as all `1'.
The number of missing conditions is calculated according to Eq.~(\ref{eq:combination}). For instance, with two dense modalities ($n{=}2$) and two sparse modalities ($m{=}2$), the expected number of missing conditions is $C{=}12$.
Since the number of missing conditions $C$ is always much less than the number of training epochs (\eg, $200$), the model can thoroughly explore all the missing conditions of all $d$ samples during training, which means the data amount is augmented to $C{\times}d$. 
Unlike the training strategy with a predetermined missing ratio that requires a minimum of $d$ Bytes to store the mapping between $d$ image indexes and $C$ missing conditions, our proposed MMS strategy utilizes only $m{+}n$ bits to implement modality dropout, demonstrating efficiency regardless of dataset size. 
According to previous works~\cite{lee2023missing_prompt, wang2023multi}, training with missing modalities often leads to a performance decrease on validation set with complete samples. 
However, our MMS ensures that the models remain robust even when the predominant modality is missing, without sacrificing their performances on complete data.

\subsection{Fourier Prompt Tuning}
\begin{figure}
    \centering
    \includegraphics[width=1.0\columnwidth]{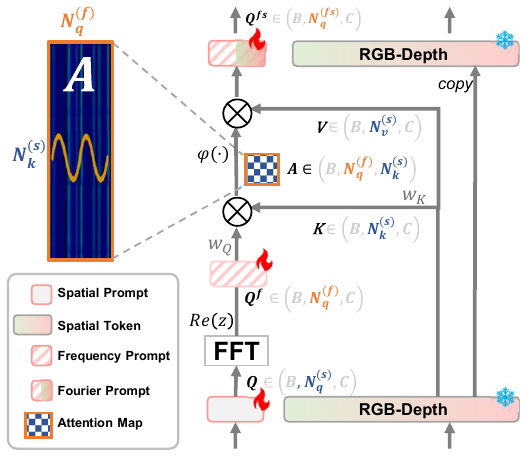}
    \vskip -1ex
    \caption{\small \textbf{Fourier Prompt Tuning} module. 
    Through Fast Fourier Transformation (FFT) and the interaction with spatial tokens (\eg, RGB-Depth), the resulting prompt, although with limited tunable parameters, contains both \textit{spectral} and \textit{spatial} information to robustify the frozen model in the modality-incomplete context.}
    \label{fig:fourier_tuning}
    \vskip -2ex
\end{figure}

Prompt tuning, a parameter-efficient tuning method, involves adding a small number of tunable prompt tokens (\eg, $200$ in this study) alongside the input or feature tokens (${\sim}5000$), with the backbone remaining frozen. This approach allows the models to efficiently adapt to downstream tasks, and the prompt tokens effectively compensate for information loss arising from incomplete modalities. Thus, we employ prompt tuning to address MISS.
Regarding the information to be injected into the prompts, we consider \textit{spatial} and \textit{spectral} information. 
Although spatial information is essential for semantic segmentation, its reliability is hindered by noise arising from missing and variable modalities, while spectral information remains globally representative.
Hence, we incorporate spectral information into prompts using Fast Fourier Transformation (FFT), rectified across all feature tokens via cross-attention, as shown in Fig.~\ref{fig:fourier_tuning}.
This seamless integration
enables effective compensation for the presence of noisy feature tokens.

The discrete Fourier Transformation (DFT) is a mathematical operation that transforms a finite sequence $\{x_n\}$ consisting of equally spaced samples into another sequence $\{X_k\}$ of the same length, which is characterized as a complex-valued function of frequency. One implementation of DFT is FFT. 
FFT and Inverse Fast Fourier Transformation (IFFT) are defined through the following formulas:
\begin{equation}
    X_k=\sum_{n=0}^{N-1}x_ne^{-\frac{2\pi i}{N}nk},\quad 0\leq k \leq N-1.\
    \label{eq:fft}
\end{equation}
\begin{equation}
    x_n = \frac{1}{N}\sum_{k=0}^{N-1}X_ke^{\frac{2\pi i}{N}nk},\quad 0\leq n\leq N-1.
\end{equation}
Eq.~(\ref{eq:fft}) shows that FFT possesses two distinct properties: the ability to extract frequencies and facilitate global interaction.
Previously, these properties were utilized for two purposes separately. 
In the works~\cite{li2023feature, guibas2021adaptive}, operations are implemented between FFT and IFFT for frequency analysis, but their combination introduces information loss due to digitalization. 
On the other hand, FNets~\cite{lee2021fnet, sevim2022fast} use only the global interaction property of FFT, replacing the self-attention mechanism and operating as a token mixer. 

In our Fourier Prompt Tuning method, as illustrated in Fig.~\ref{fig:fourier_tuning}, 
we take advantage of
both essential FFT properties for enhancing MISS.
As prompt tokens and feature tokens are processed together by transformer layers and mixed there, we should separate the tokens $E_l$ into prompt tokens $P_l$, feature tokens $Z_l$ and a classification token $Z_{cls}$:
\begin{equation}
	P_{l}, [Z_l, Z_{cls}] = \mathrm{split}\left(E_{l}\right),
\end{equation}
where $l$ represents the number of layers where our Fourier Prompt module is implemented.
For better understanding, we write the method in the format of cross-attention: 
\begin{equation}
    Q_{l} = P_l,\quad K_l=V_l=Z_l,
\end{equation}
so the number of prompt tokens is $N_q$ and the number of feature tokens is $N_k, N_v$. 
We use FFT to calculate the spectrum of the prompt $Q_l$ and just take the real part of the complex numbers by using a $\textbf{Re}{(\cdot)}$ operator:
\begin{equation}
    Q_l^f = \textbf{Re}(\textbf{FFT}(Q_l)).
\end{equation}
The superscript represents the information involved in the tokens of frequency ($f$) and spatiality ($s$), respectively. 
By default, the tokens include spatial information. Then, the spectrum of the prompt is rectified by all feature tokens through cross-attention: 
\begin{equation}
	Q_{l}^{fs} = softmax\left(\frac{(W_{Q,l}{\cdot}Q_{l}^{f})(W_{K,l} {\cdot} K_{l})^T}{\sqrt{d}}\right)V_{l},
\end{equation}
where $d$ denotes the number of channels. The weight matrices for queries $W_{Q,l}$ and keys $W_{K,l}$ can be regarded as channel-mixers. Feature tokens $V_l$ (dimension $(B, N_v^s, C)$) multiplied with the attention matrix (dimension $(B, N_q^f, N_k^s)$) facilitate the restoration of the original spatial information.
The module's output is the prompt $Q_l^{fs}$ with spectral information rectified by all feature tokens. 
The last step is to replace $P_l$ with our Fourier Prompt $P_l^{fs}$:
\begin{equation}
    P_l^{fs}=Q_l^{fs},
\end{equation}
\begin{equation}
	E^{fs}_{l} =\mathrm{concat}[P_{l}^{fs}, Z_l, Z_{cls}].
\end{equation}
As the transformers have hundreds of channels, the weight matrices are heavy. 
To harness the benefits of adapters~\cite{chen2022adaptformer} and minimize the number of parameters, 
we integrate our Fourier Prompt module into the AdaptFormer framework. Consistent with~\cite{patro2023spectformer}, this integration occurs within the down-up bottleneck of the initial four blocks out of twelve. Subsequently, for parameter efficiency, we replace FPT with a cross-attention layer without learnable parameters in the remaining blocks. 

\section{Experiments}
\label{experiments}

\input{tex/setup}
\input{tables/deliver}

%
\subsection{Comparison with the State of the Art}

\noindent\textbf{Baseline of the MISS task.} 
We compare our methods 
with prompt tuning methods~\cite{jia2022vpt, yoo2023gvpt, lee2023missing_prompt} and the \textbf{AdaptFormer} framework~\cite{chen2022adaptformer}.
Visual Prompt Tuning (\textbf{VPT})~\cite{jia2022vpt} is the vanilla prompt tuning method that adds several learnable tokens alongside only the input tokens (Shallow) or feature tokens of each layer (Deep). 
Gated Prompt Tuning (\textbf{Gated VPT})~\cite{yoo2023gvpt} identifies variations in the optimal prompt token layer for self-supervised and supervised pre-trained models. %
Since our backbone is pre-trained in a self-supervised fashion with MultiMAE~\cite{bachmann2022multimae}, we opt to adopt Gated VPT as our baseline.
The pioneering work of Missing-aware Prompt Tuning (\textbf{Missing-P})~\cite{lee2023missing_prompt} introduces prompt tuning to tasks involving missing modalities, where $C$ sets of prompts align with $C$ distinct missing conditions. 

\noindent\textbf{Effectiveness of Fourier Prompt Tuning (FPT).}
\input{tables/cityscapes}
We assess the performance of our Fourier Prompt Tuning (FPT) on the DeLiVER~\cite{zhang2023delivering} and Cityscapes~\cite{cordts2016cityscapes} datasets with different scenarios, including MISS conditions, sensor failures, system failures, and normal situations. On each dataset, we further conduct two groups of experiments, \ie, training with and without modality-missing strategies.

Tab.~\ref{tab:deliver} shows two groups of results on DeLiVER. %
None of the baseline models achieve significant improvement over decoder tuning in both missing conditions. However, our robust FPT demonstrates considerable gains, achieving $7.32\%$ and $0.89\%$ increases in mIoU over decoder tuning when RGB and Depth are respectively absent.
The observed $1.97\%$ mIoU improvement over AdaptFormer~\cite{chen2022adaptformer} underscores the efficacy of our FPT in addressing %
sensor failures of the DeLiVER dataset. 
When training with modality dropout strategies, the training strategy employed by Missing-P~\cite{lee2023missing_prompt} is distinctively its own, while the remaining approaches are trained with our MMS.
Although Missing-P is proficient in scenarios lacking RGB, it increases dependency on the Depth modality due to the predefined and imbalanced missing ratio. Conversely, our FPT, with an approximate missing ratio of $50\%$ for two dense modalities, outperforms Missing-P, which has a missing ratio of $70\%$, across all conditions.
Our FPT consistently surpasses AdaptFormer~\cite{chen2022adaptformer}, by $1.64\%$, $0.27\%$, and $2.84\%$ mIoU in the cases of RGB-Depth, RGB missing, and Depth missing, respectively.

Tab.~\ref{tab:cityscapes} shows two groups of results on Cityscapes, where our analysis exclusively focuses on system failures, \textit{i.e.} missing modalities. 
When trained on full RGB-Depth data, our FPT significantly outperforms AdaptFormer, with a $3.56\%$ mIoU increase. Under the MSS strategy, FPT shows even greater improvements over AdaptFormer in various scenarios: $3.8\%$ for full RGB-Depth, $5.84\%$ with RGB missing, and $2.78\%$ with Depth missing.
\input{tables/two_modalities}
\input{tables/four_modalities}

\noindent\textbf{Effectiveness of Missing-aware Modal Switch (MMS).} 
Previously, the effectiveness of our MMS was established in parameter-efficient tuning methods with a frozen backbone. 
Now, we delve into the influence of MMS specifically within the domain of full fine-tuning. 
In Tab.~\ref{tab:2_modal}, we compare the performances of our MMS and the approach proposed by Lee~\etal~\cite{lee2023missing_prompt} on a one-stream model (MultiMAE) and a multi-stream model (CMNeXt) on DeLiVER and Cityscapes datasets. 
The training strategy in~\cite{lee2023missing_prompt} is to randomly split the dataset with a missing ratio of $70\%$. 
According to~\cite{lee2023missing_prompt, wang2023multi}, an increase in the missing ratio leads to degraded performance on complete validation sets and may occasionally intensify the reliance on predominant modalities.
Our MMS acts as a data augmentation method, improving model performance in scenarios with missing modalities on both datasets, especially when predominant modalities are absent.
Notably, on Cityscapes, MMS boosts models by up to $47.5\%$ in mIoU when RGB, the predominant modality, is absent.
Additionally, there is no compromise in performances observed on complete sample pairs, with a variance within the range of $\pm0.5\%$ in mIoU.

\begin{figure}
    \centering
    \includegraphics[width=.9\columnwidth]{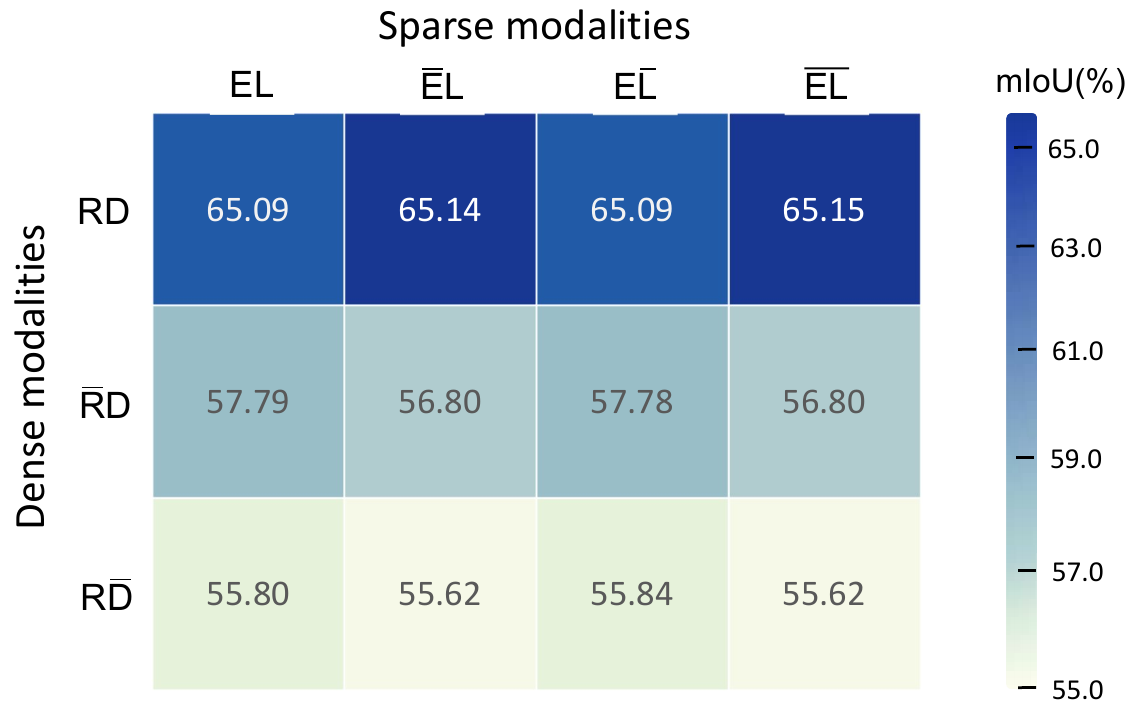}
    \vskip -1ex
    \caption{\textbf{Results of different combinations of missing modalities}, including dense (RGB and Depth) and/or sparse (Event and LiDAR) modalities. The overline means modality missing, \eg $\overline{\mathrm{R}}$.
    }
    \label{fig:results_combinations}
    \vskip -4ex
\end{figure}

Given that MMS is specifically tailored to handle dense and sparse modalities differently, the importance of training with the absence of sparse modalities is explored in Tab.~\ref{tab:4_modal}. Note that the conducted experiments adhere to the initial configuration as outlined in CMNeXt~\cite{zhang2023delivering}. 
Specifically, the depth maps are processed using the HHA method, and the images are resized to dimensions of $1024{\times}1024$. 
When trained with MMS integrating both dense and sparse modalities, CMNeXt outperforms the one trained exclusively with MMS of dense modalities in scenarios involving a single missing modality.
Fig.~\ref{fig:results_combinations} illustrates the outcomes across all $12$ instances of missing data while keeping at least one dense modality trained with MMS. 
There is no performance decrement when the model is trained using our MMS in scenarios where only sparse modalities are missing. This observation substantiates the premise that sparse modalities bolster the performance of dense modalities in the semantic segmentation task.

\input{tables/Ablation}
\input{tables/conditions}
\subsection{Ablation Studies}
\noindent\textbf{Analysis of prompt spaces.} 
We apply Fast Fourier Transformation (FFT) to the feature tokens in order to investigate only the spectrum space while excluding FFT from our FPT to concentrate on the spatial characteristics. 
As shown in Tab.~\ref{tab:ablation_study}, integrating spatial information improves performance in scenarios with missing modalities. 
Conversely, tuning prompts in the spectral domain enhances the performance on complete data. 
The synergy between both domains yields optimal performances in complete and missing conditions.

\noindent\textbf{Analysis in comprehensive MISS cases.}
In Tab.~\ref{tab:res_condition}, we compare our FPT with two representative parameter-efficient counterparts trained with and without our MMS. 
The performances of the methods
in occurrences of five distinct weather conditions and five instances of sensor failures from the DeLiVER benchmark~\cite{zhang2023delivering} are enumerated. 
On average, methods trained with MMS significantly outperform those trained with complete sample pairs in scenarios of \textit{night} and \textit{under-exposure}.
However, 
models trained with complete sample pairs 
exceed those using our MMS in other sensor failure cases, because the DeLiVER dataset inherently has diverse adverse conditions in both training and validation sets, acting as effective data augmentation. 
With the minimum tunable parameters, our FPT achieves state-of-the-art in all conditions. Notably, in scenarios like \textit{night}, \textit{motion blur}, \textit{over-exposure}, \textit{under-exposure}, and \textit{LiDAR-jitter}, our FPT outperforms the second-best method by over $2\%$ in mIoU.

\begin{figure}
    \centering
    \includegraphics[width=0.9\columnwidth]{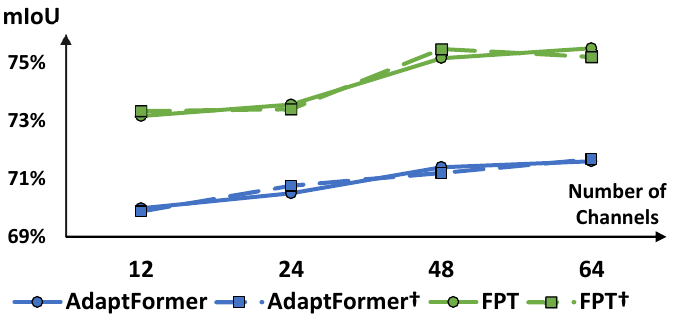}
    \vskip -1ex
    \caption{\textbf{Analysis of channels} in bottlenecks. $\dag$ is with MMS.}
    \label{fig:impact_compressed_channels}
    \vskip -5ex
\end{figure}

\noindent\textbf{Analysis of channels in bottlenecks.}
AdaptFormer~\cite{chen2022adaptformer} sets the number of channels to $64$ for optimal performance. 
In our FPT, to reduce parameters, we set the number of channels to $48$. For a fair comparison, we evaluate the performances of our FPT and AdaptFormer, trained with and without our MMS on Cityscapes, as depicted in Fig.~\ref{fig:impact_compressed_channels}. The results align with the findings in~\cite{chen2022adaptformer}, indicating that performance increases with the number of hidden dimensions. Compared to previous modality dropout strategies~\cite{lee2023missing_prompt, wang2023multi}, which result in a performance decrease when evaluated on complete samples, our MMS does not compromise the performance of either method across all hidden dimensions.

\begin{figure*}[t!]
\small
\setlength\tabcolsep{1pt}
\centering

\begin{tabular}{c c c c c c c c c c}
\rotatebox[origin=c]{90}{Depth}&
\raisebox{-0.5\height}{\includegraphics[width=0.115\textwidth]{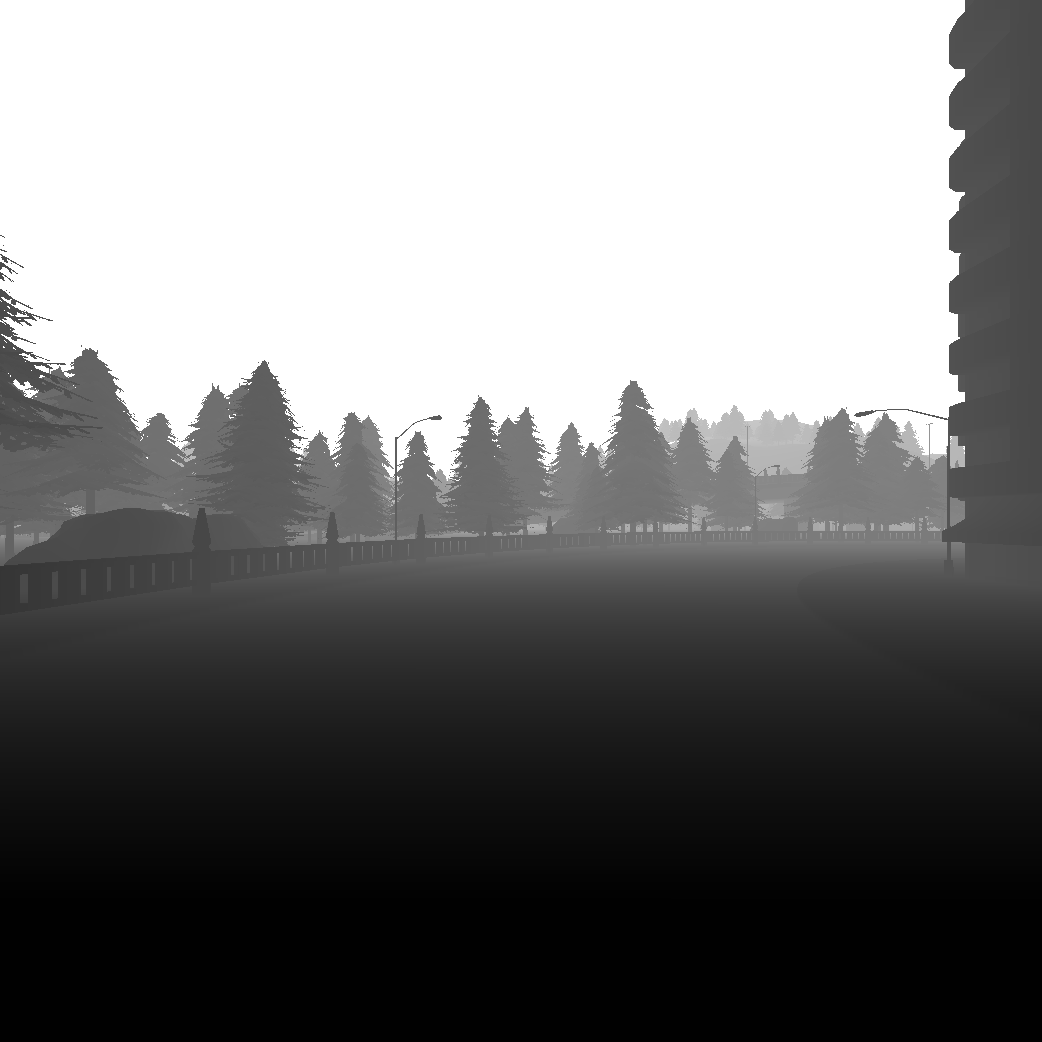}}&
\rotatebox[origin=c]{90}{\textcolor{gray!40}{RGB}-D} &
\raisebox{-0.5\height}{\includegraphics[width=0.115\textwidth]{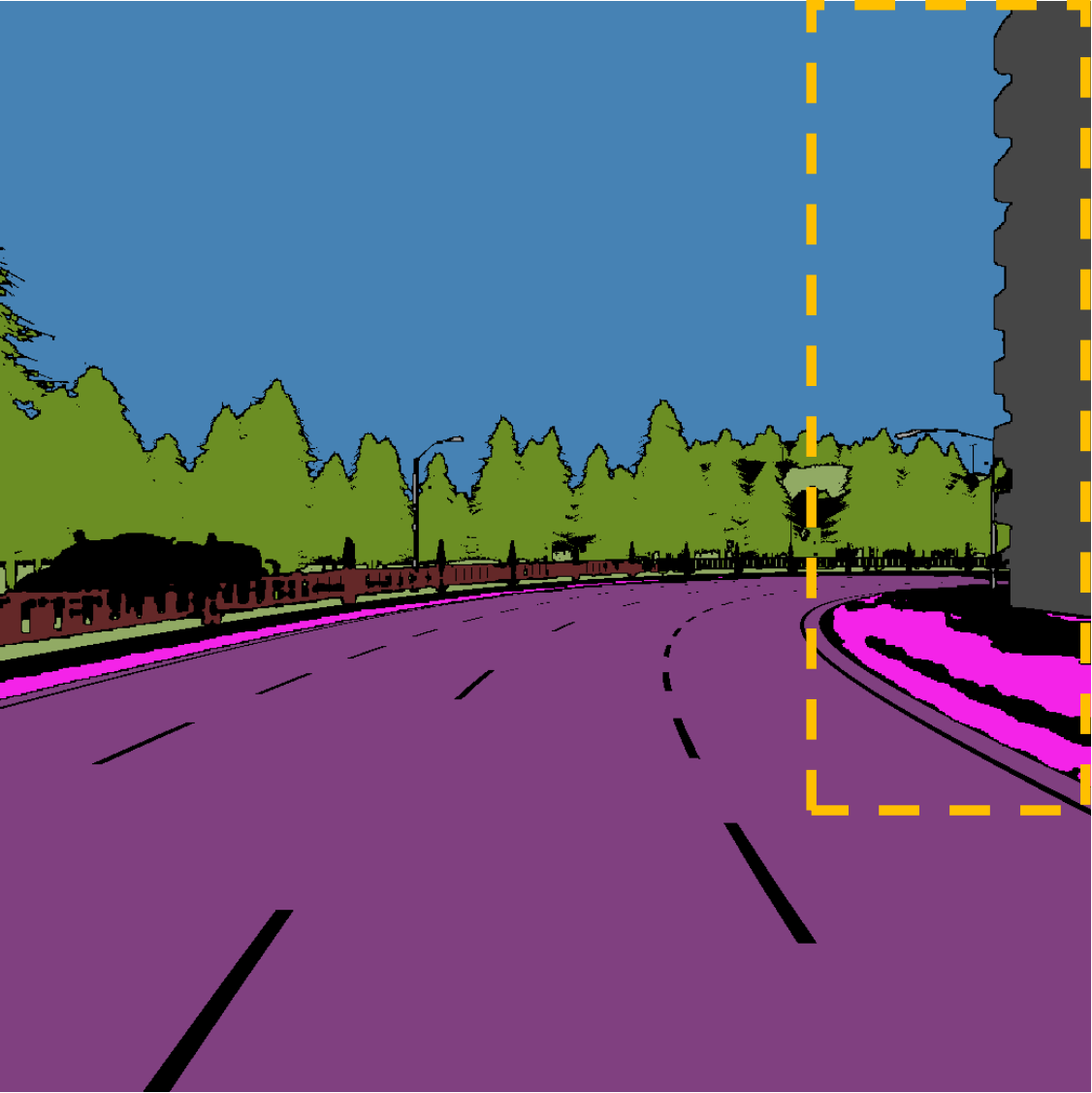}}&
\raisebox{-0.5\height}{\includegraphics[width=0.115\textwidth]{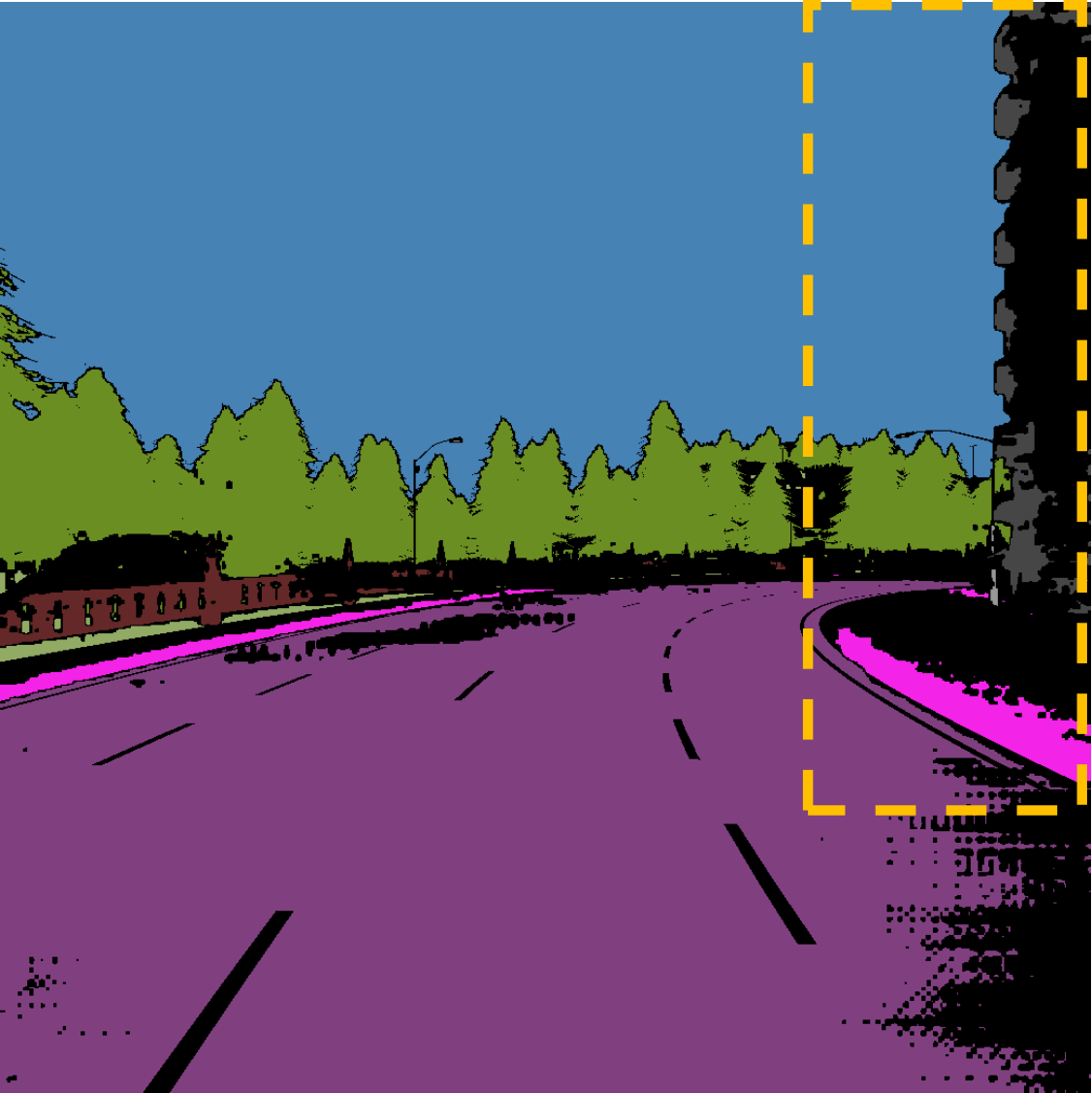}} &
\raisebox{-0.5\height}{\includegraphics[width=0.115\textwidth]{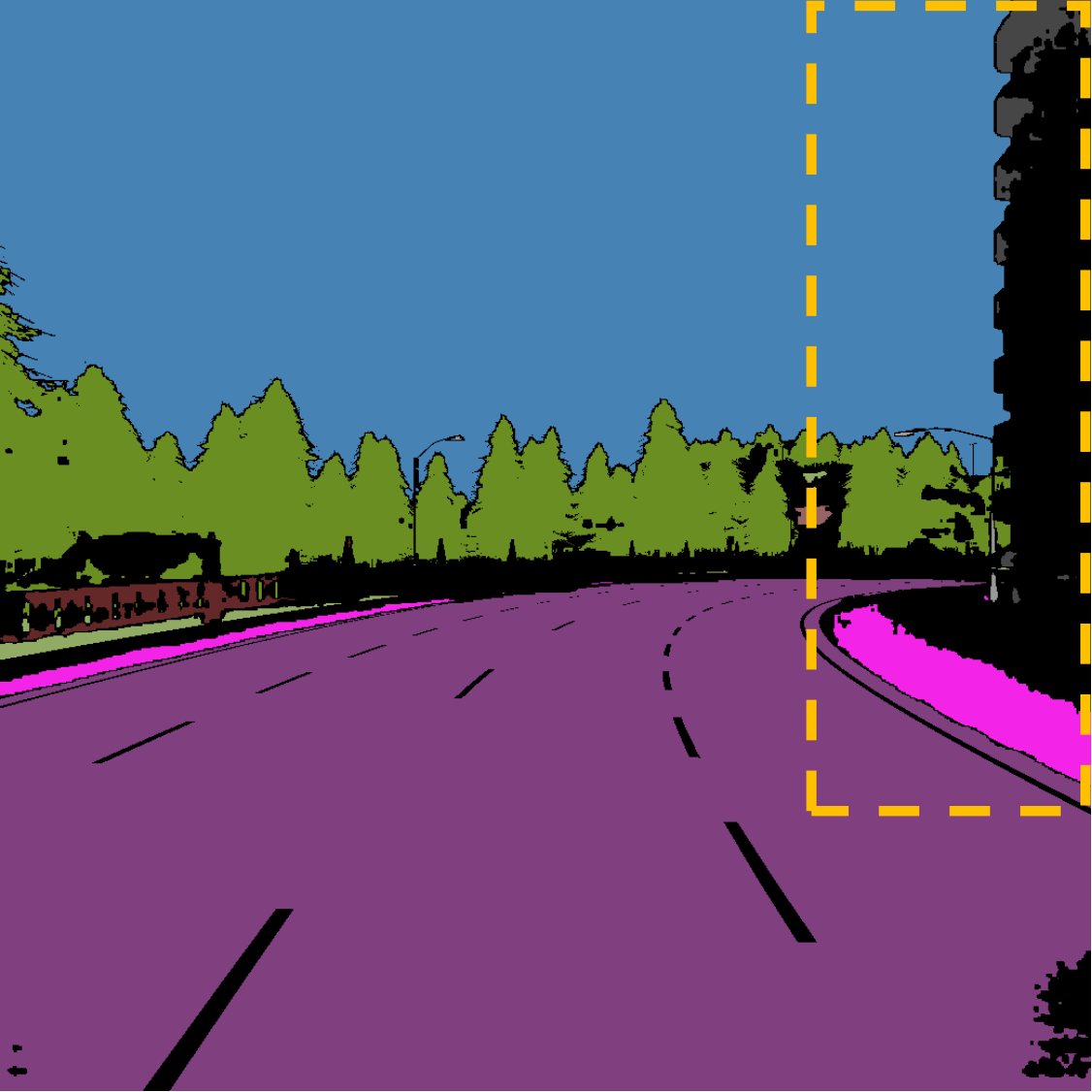}} &
\raisebox{-0.5\height}{\includegraphics[width=0.115\textwidth]{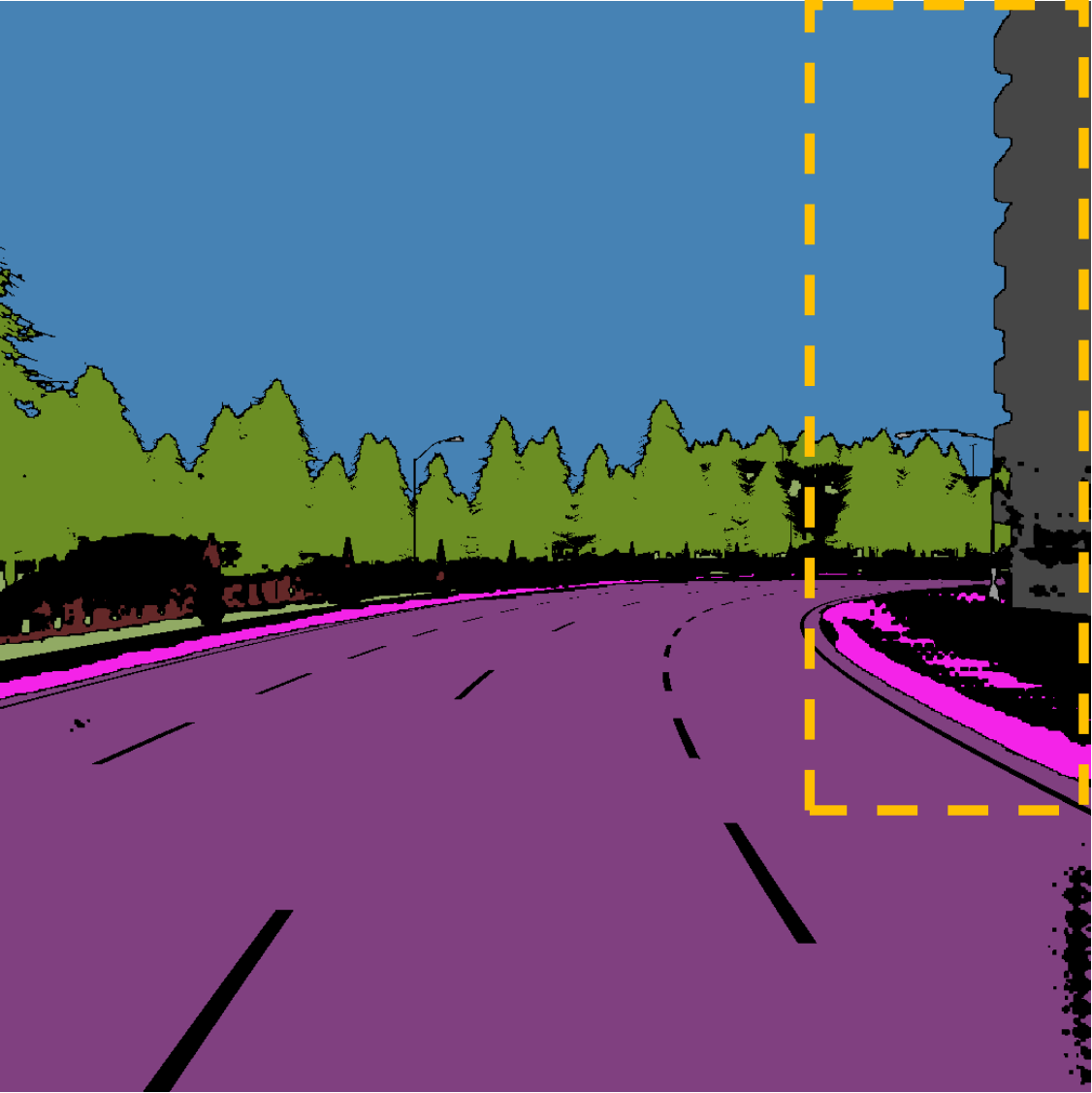}} &
\raisebox{-0.5\height}{\includegraphics[width=0.115\textwidth]{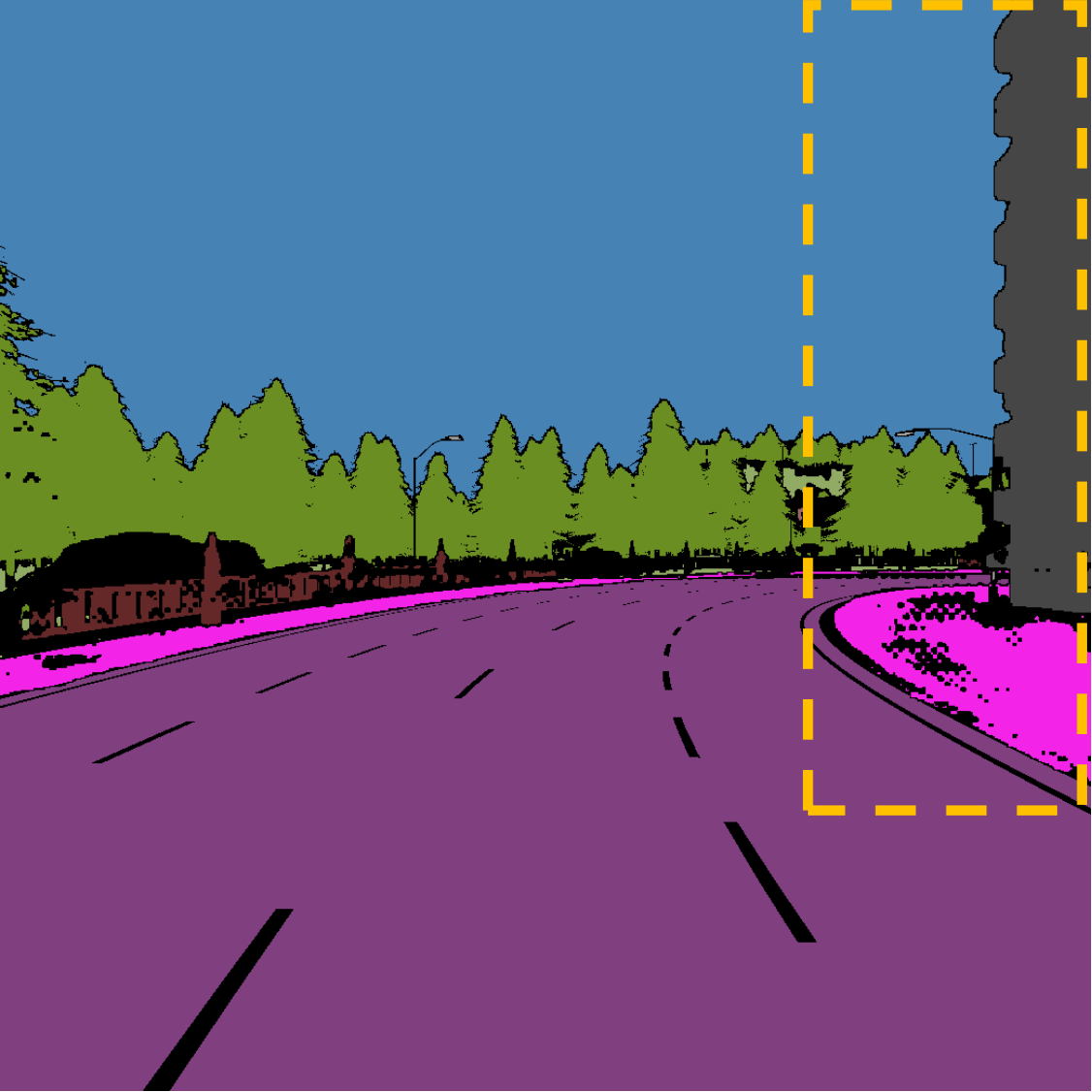}} &
\raisebox{-0.5\height}{\includegraphics[width=0.115\textwidth]{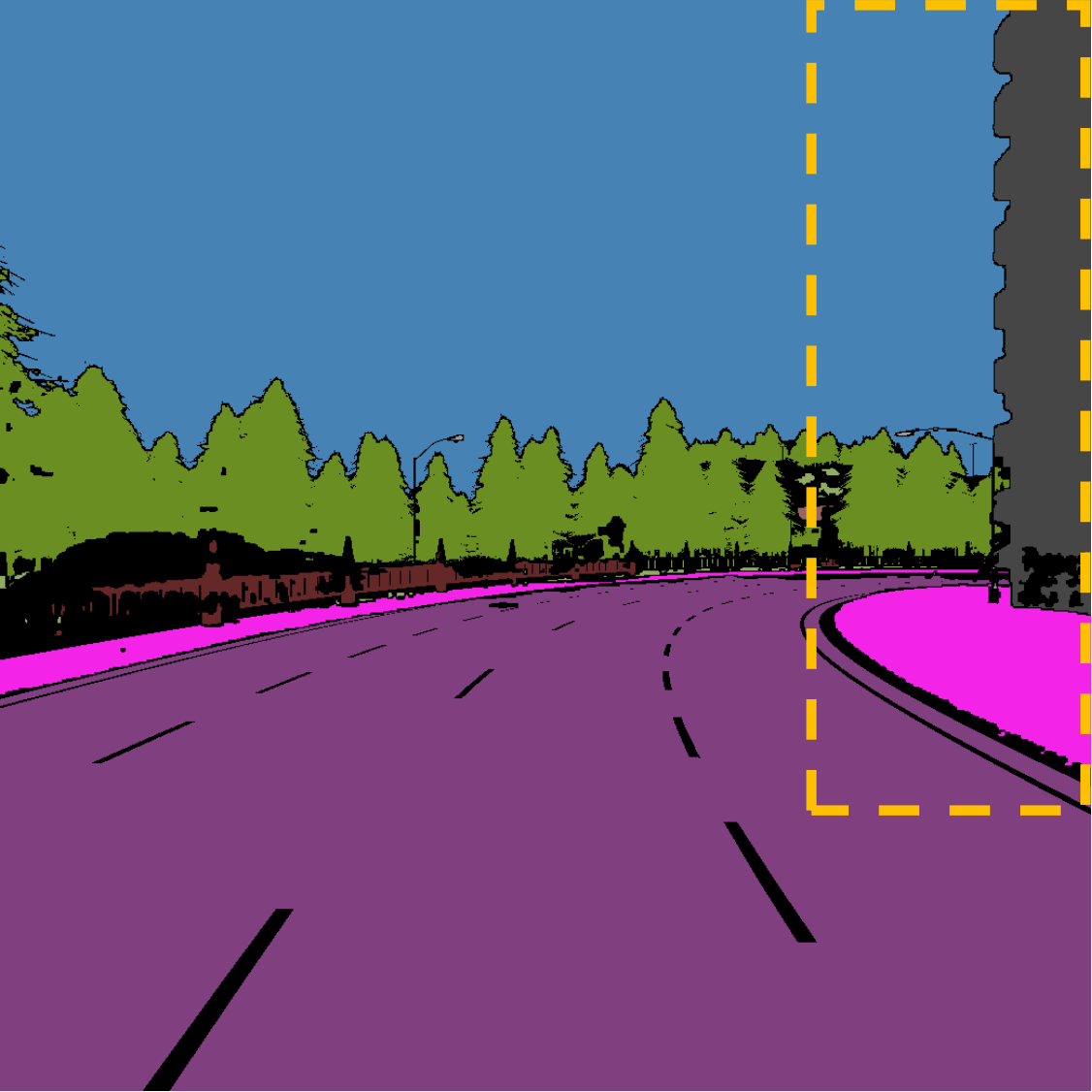}} &
\raisebox{-0.5\height}{\includegraphics[width=0.115\textwidth]{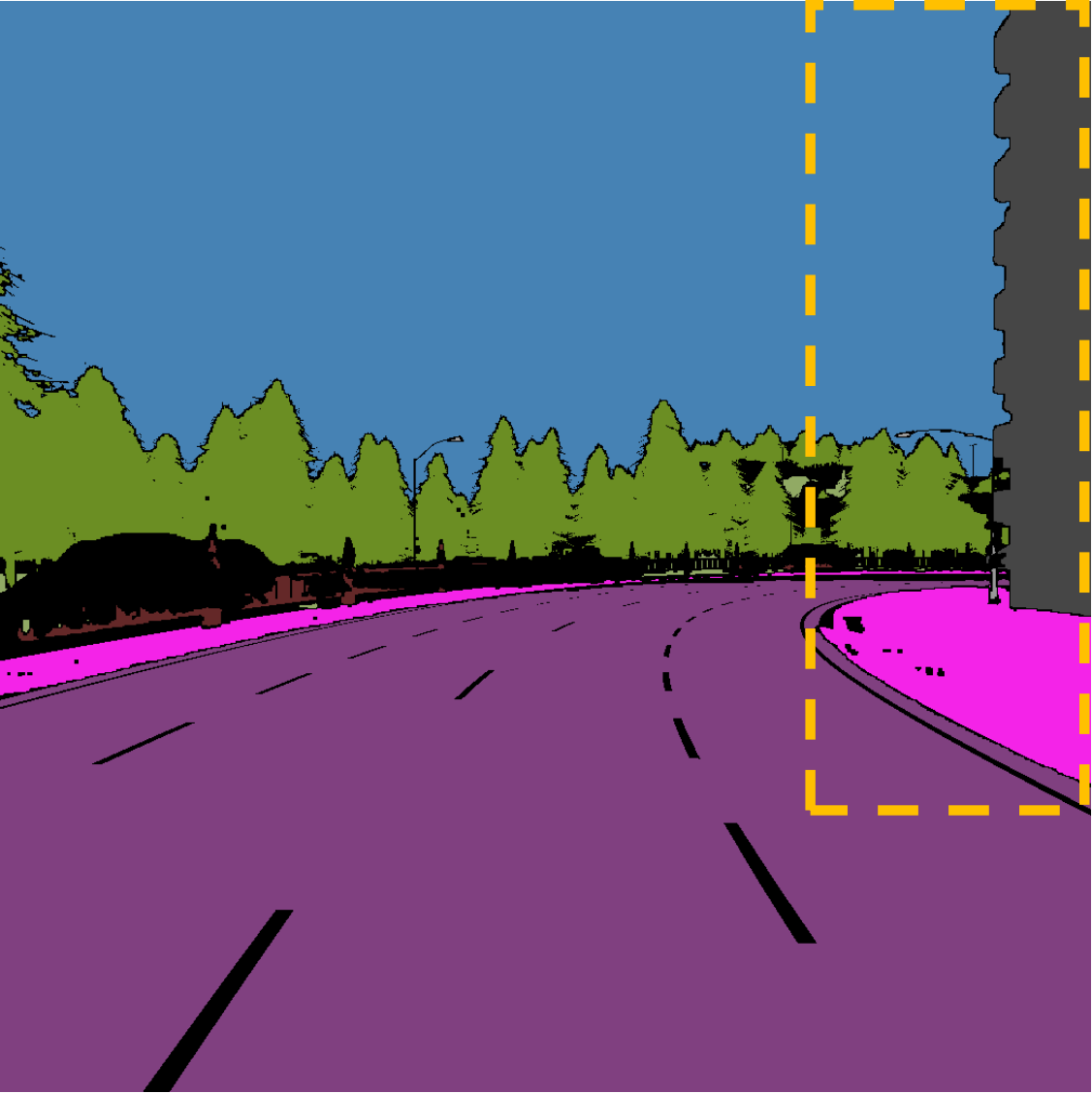}}\\
\rotatebox[origin=c]{90}{RGB}&
\raisebox{-0.5\height}{\includegraphics[width=0.115\textwidth]{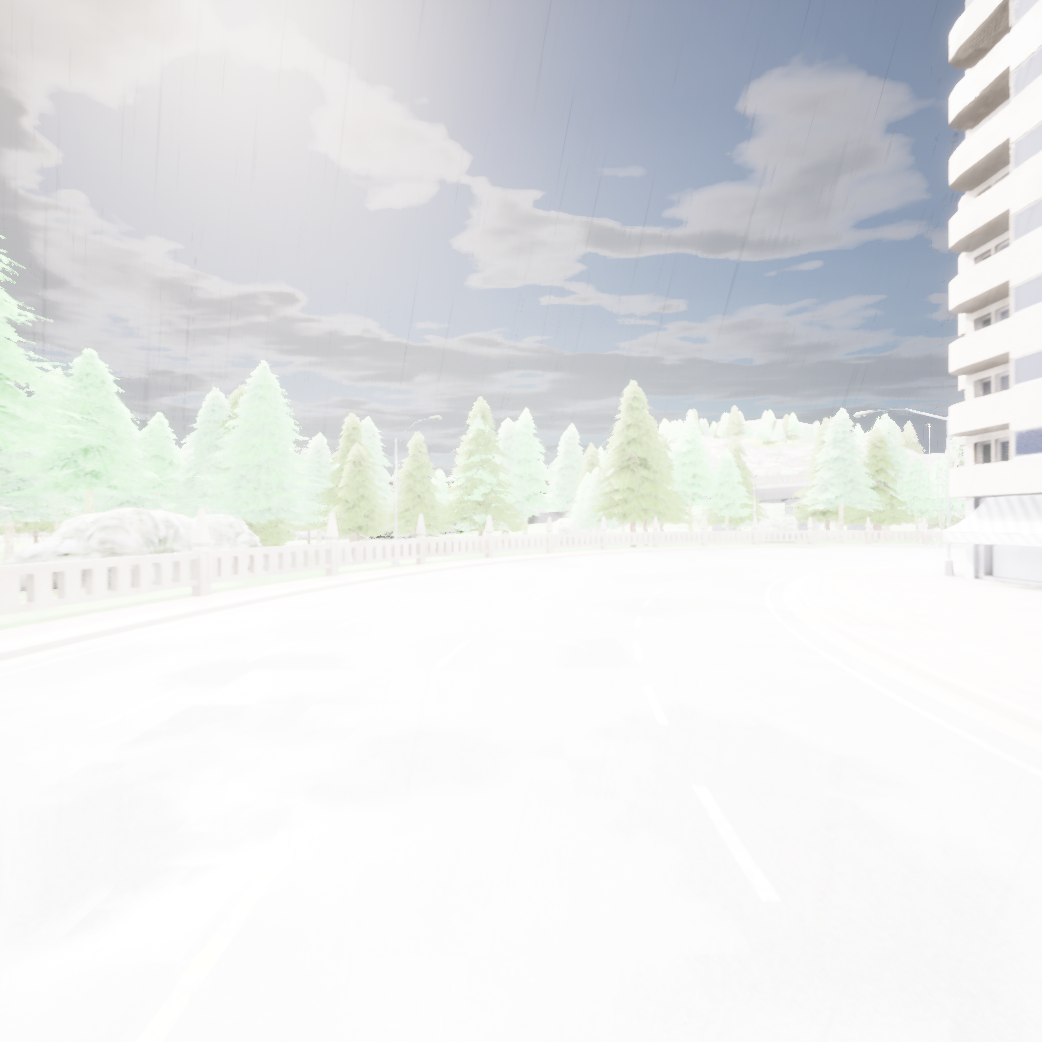}}&
\rotatebox[origin=c]{90}{RGB-\textcolor{gray!40}{D}} &
\raisebox{-0.5\height}{\includegraphics[width=0.115\textwidth]{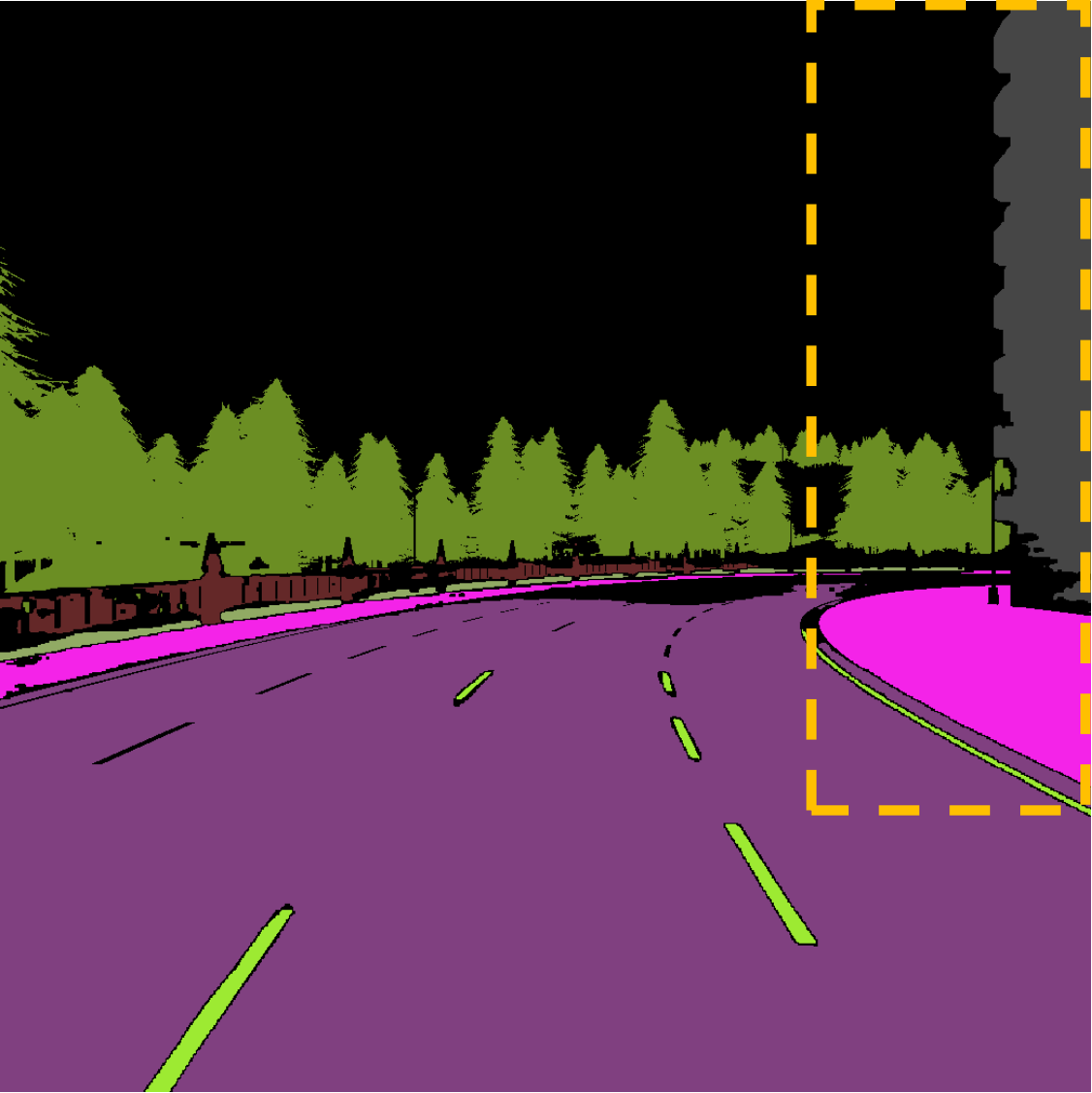}}&
\raisebox{-0.5\height}{\includegraphics[width=0.115\textwidth]{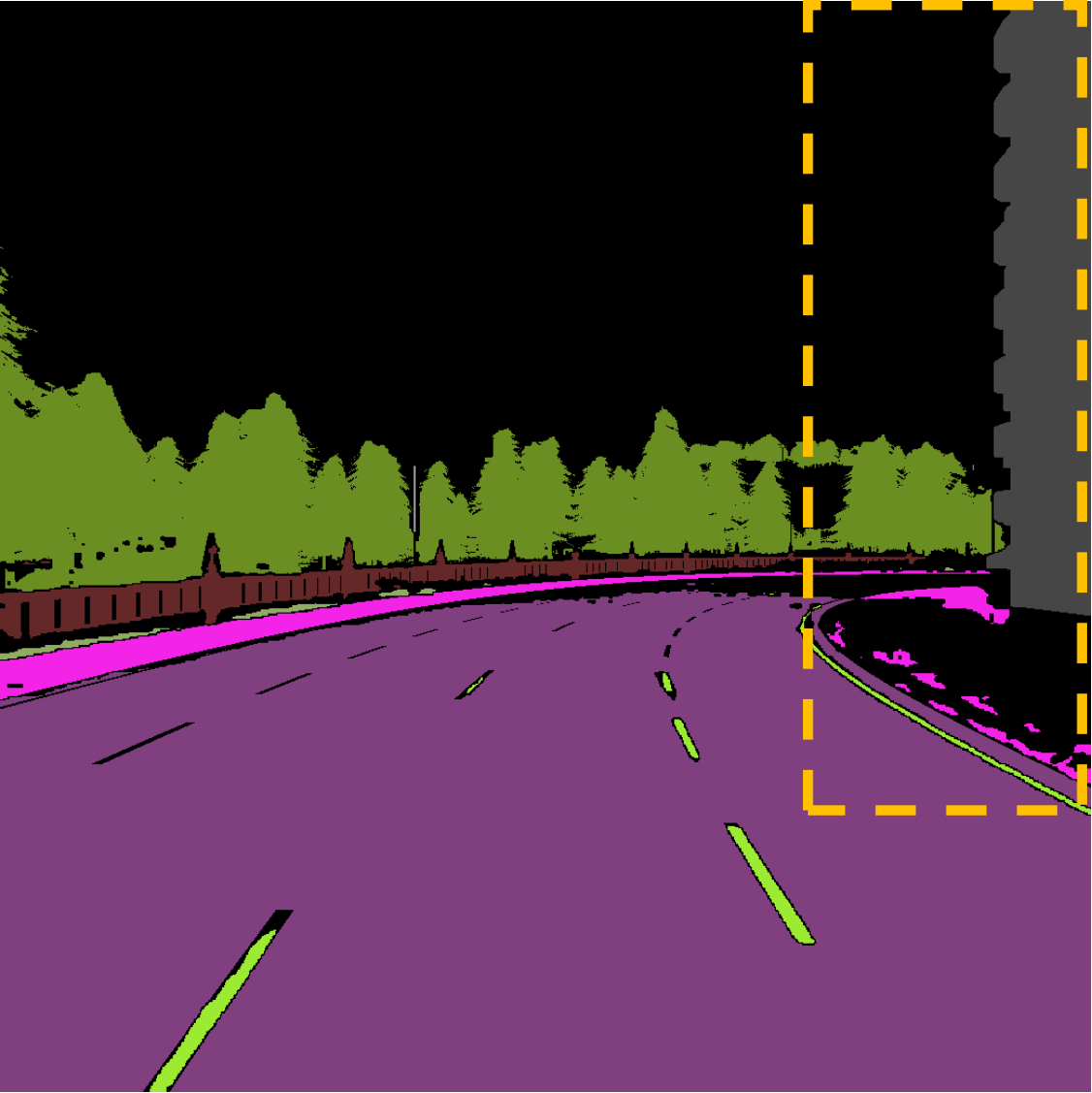}} &
\raisebox{-0.5\height}{\includegraphics[width=0.115\textwidth]{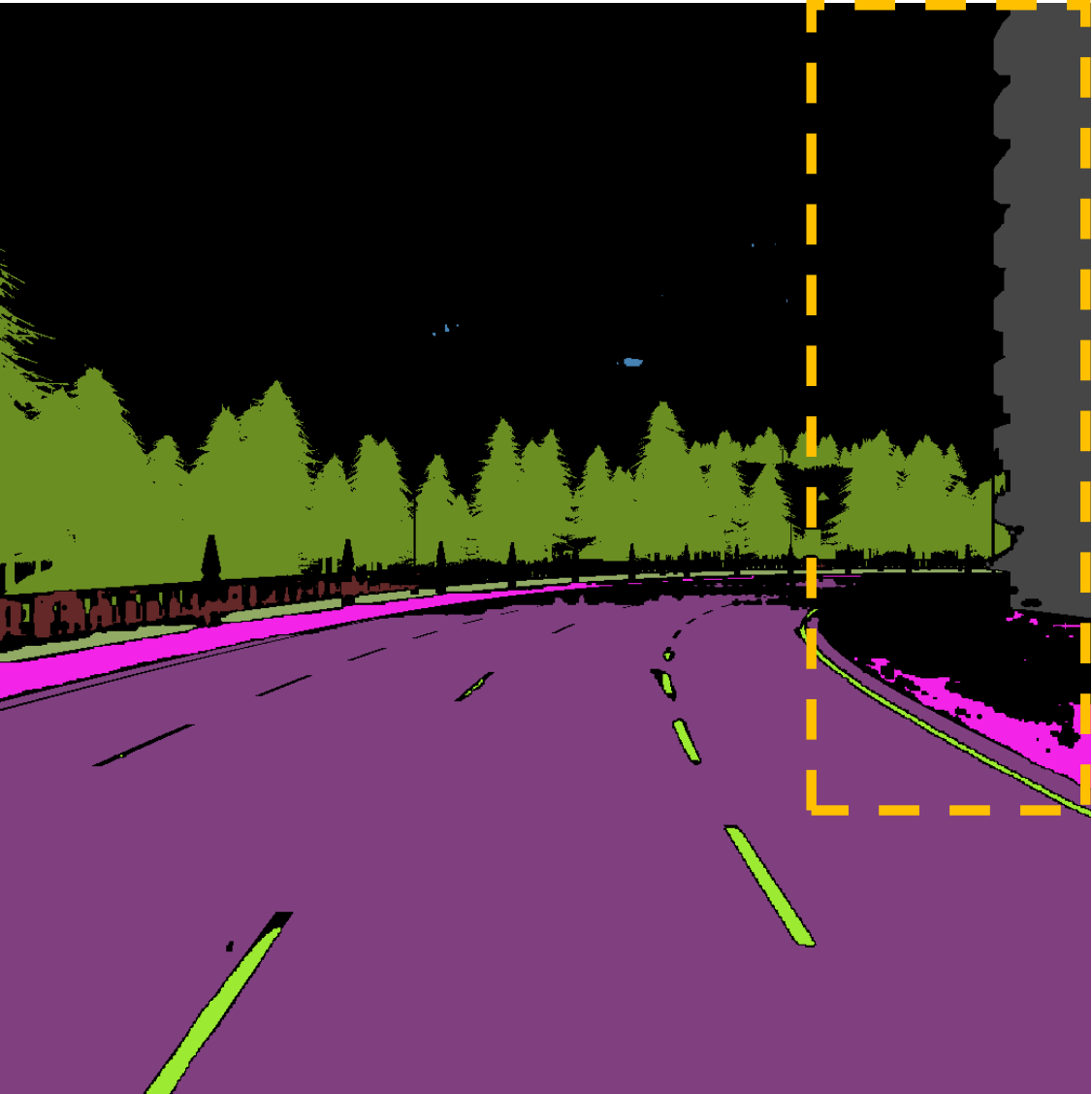}} &
\raisebox{-0.5\height}{\includegraphics[width=0.115\textwidth]{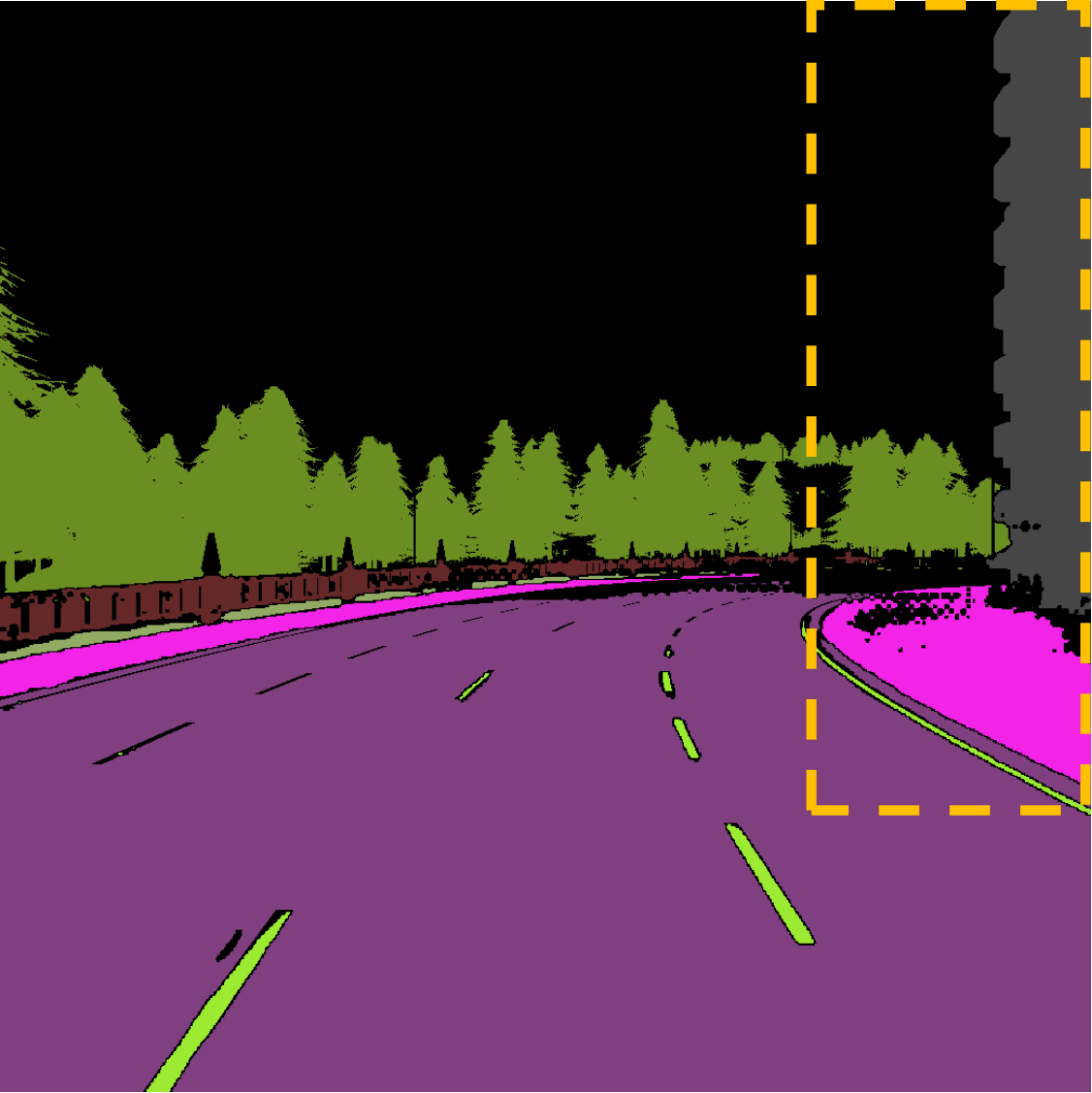}} &
\raisebox{-0.5\height}{\includegraphics[width=0.115\textwidth]{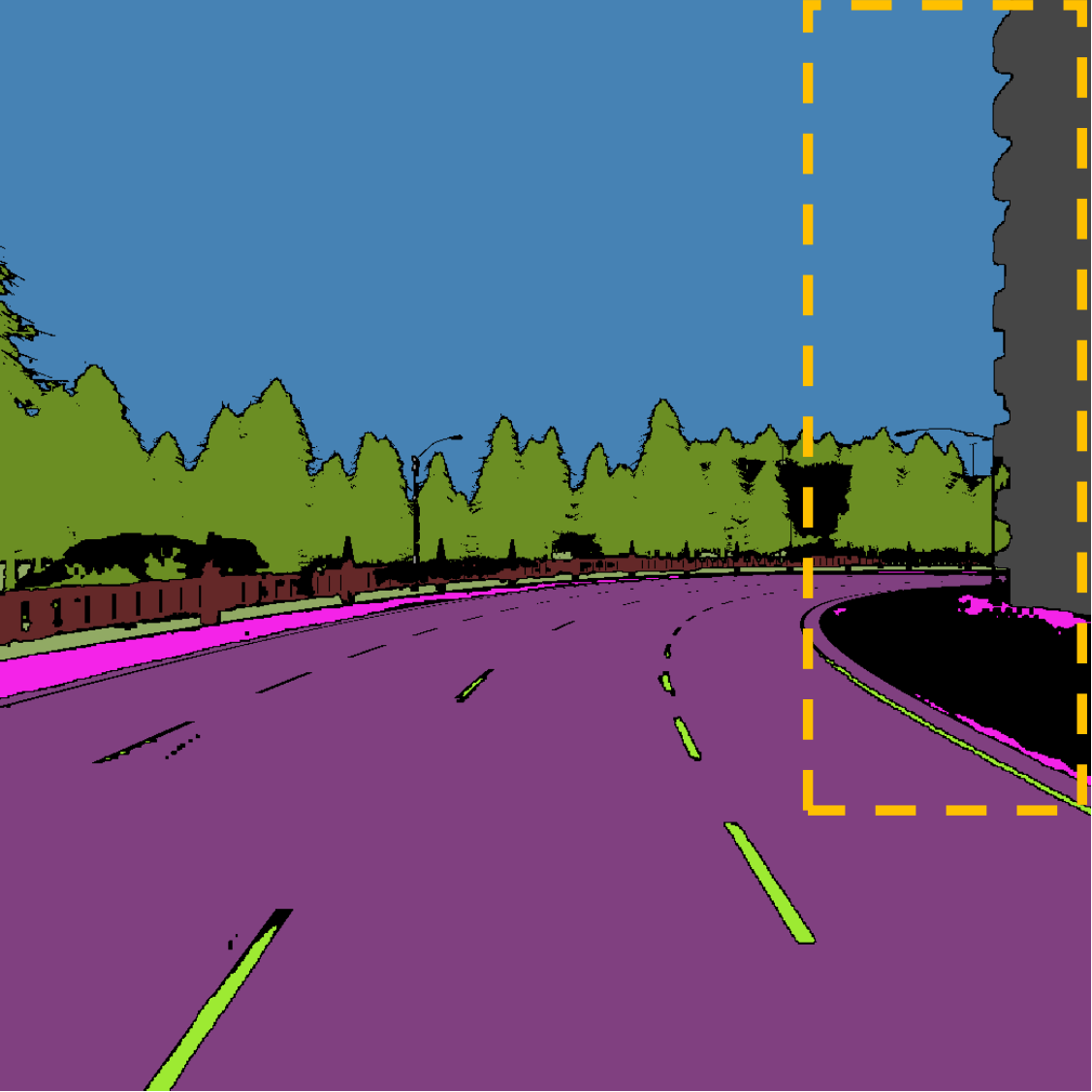}} &
\raisebox{-0.5\height}{\includegraphics[width=0.115\textwidth]{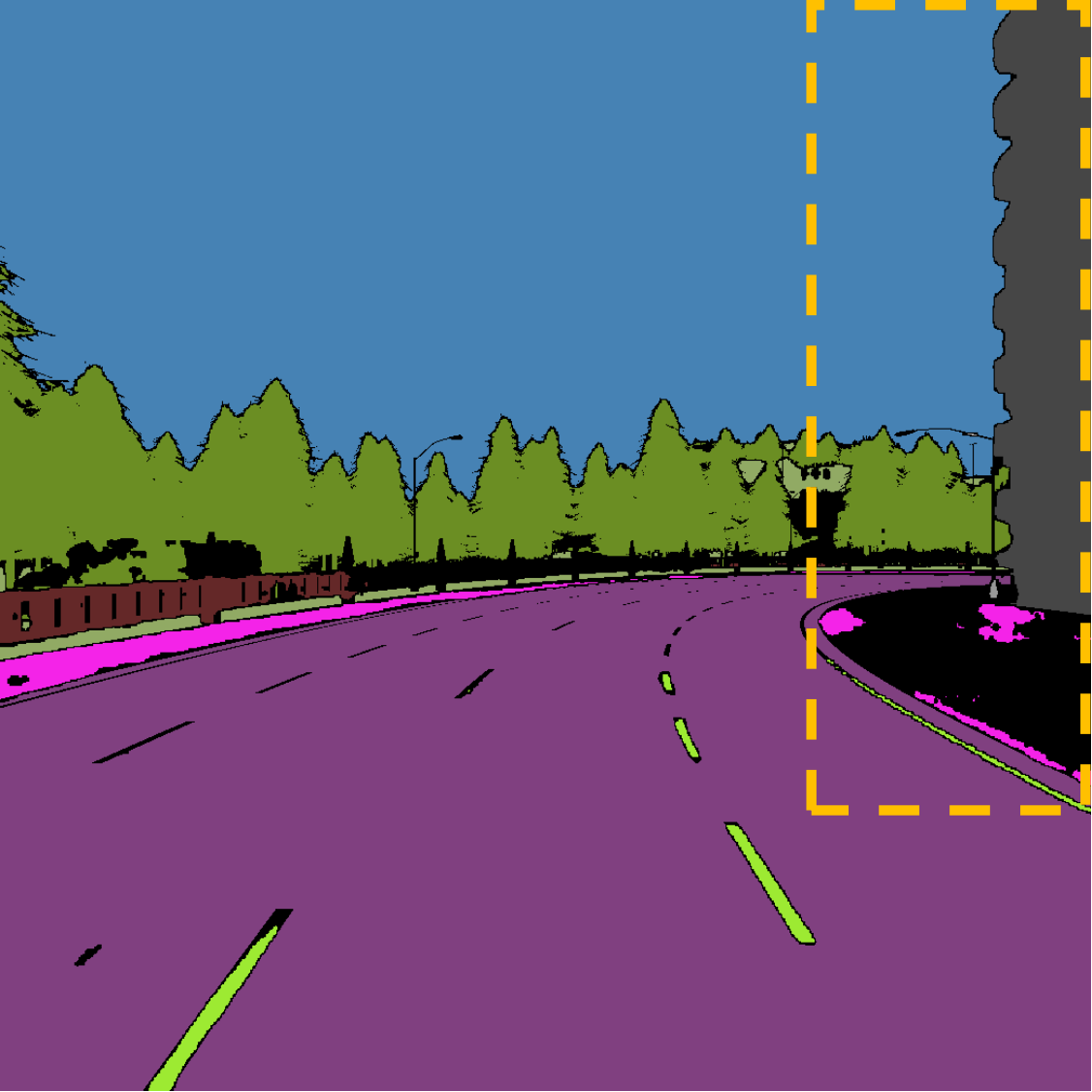}} &
\raisebox{-0.5\height}{\includegraphics[width=0.115\textwidth]{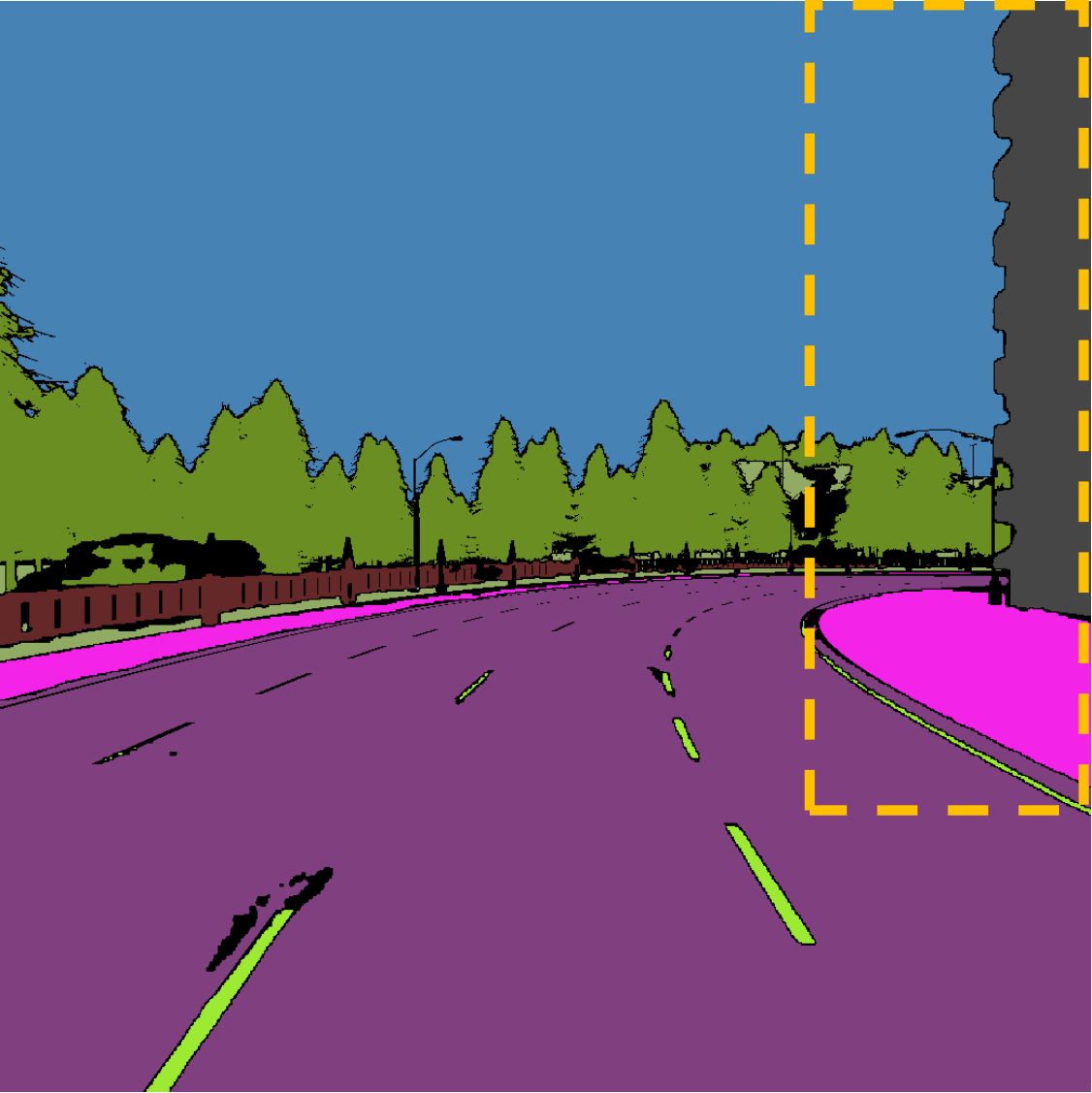}}\\
\rotatebox[origin=c]{90}{Depth}&
\raisebox{-0.5\height}{\includegraphics[width=0.115\textwidth]{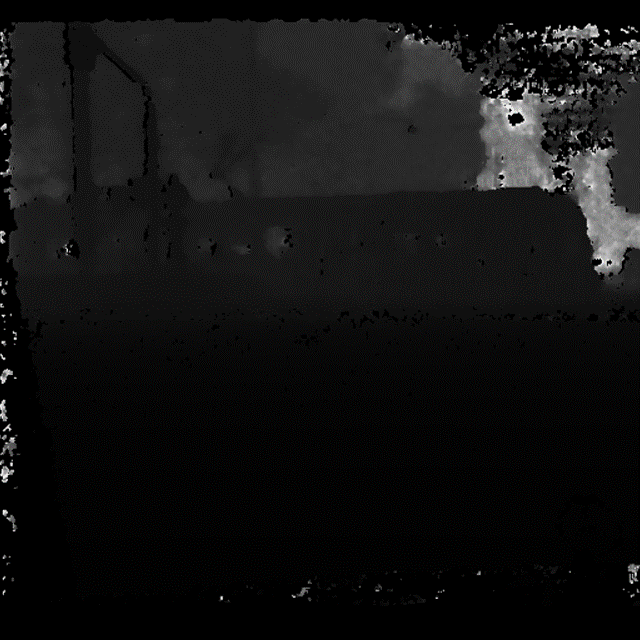}}&
\rotatebox[origin=c]{90}{\textcolor{gray!40}{RGB}-D} &
\raisebox{-0.5\height}{\includegraphics[width=0.115\textwidth]{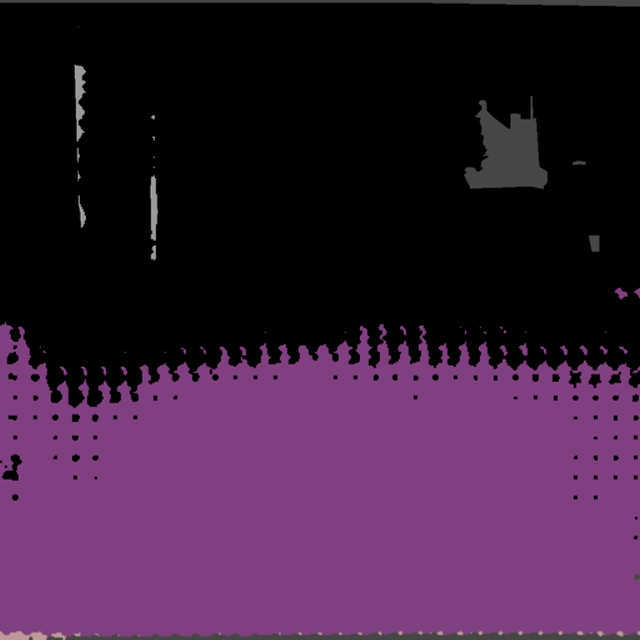}}&
\raisebox{-0.5\height}{\includegraphics[width=0.115\textwidth]{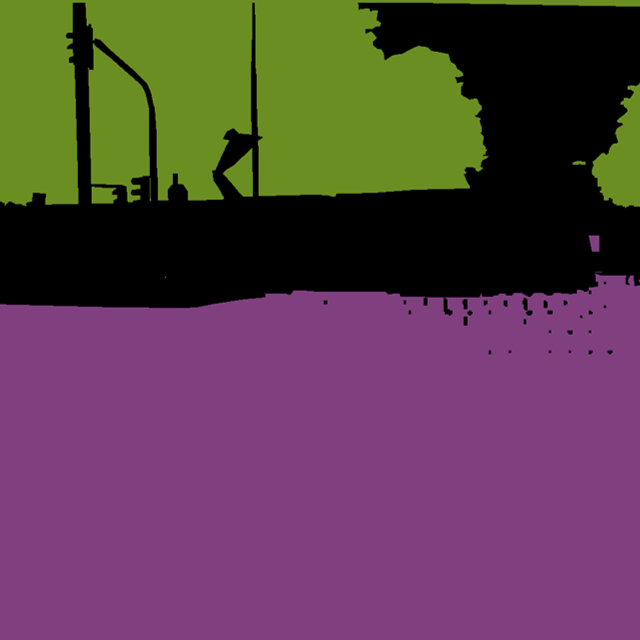}} &
\raisebox{-0.5\height}{\includegraphics[width=0.115\textwidth]{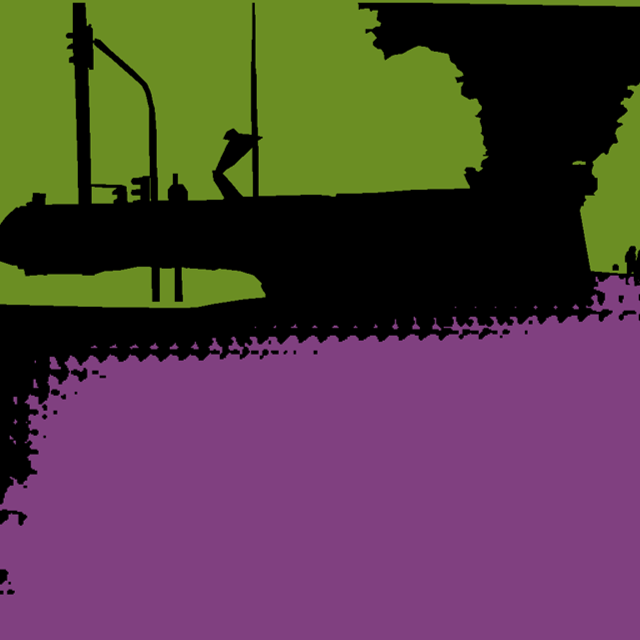}} &
\raisebox{-0.5\height}{\includegraphics[width=0.115\textwidth]{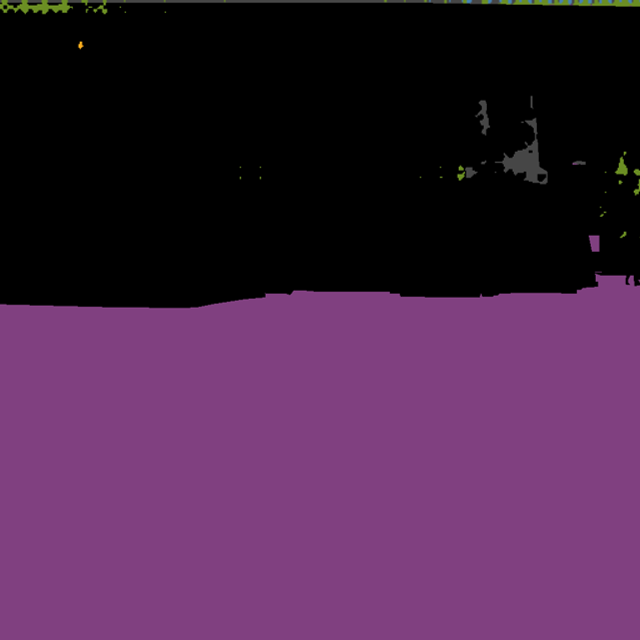}} &
\raisebox{-0.5\height}{\includegraphics[width=0.115\textwidth]{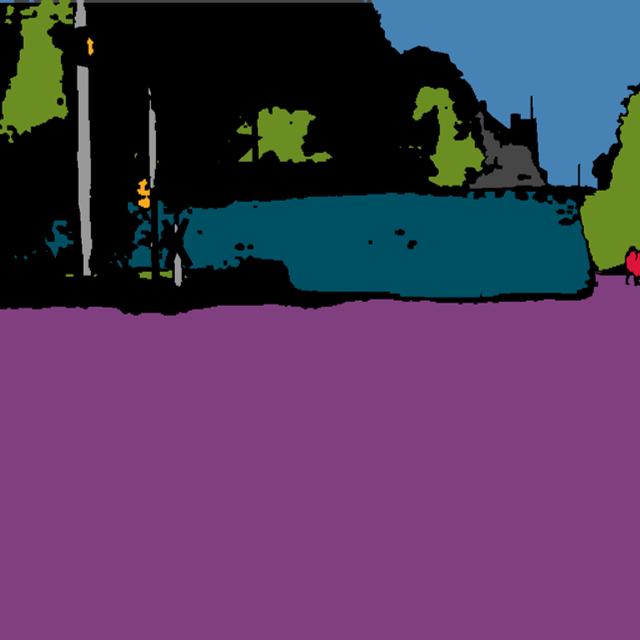}} &
\raisebox{-0.5\height}{\includegraphics[width=0.115\textwidth]{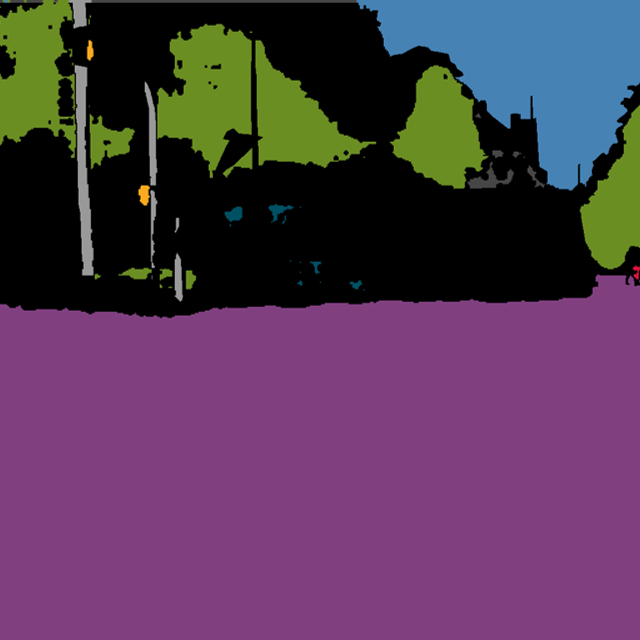}} &
\raisebox{-0.5\height}{\includegraphics[width=0.115\textwidth]{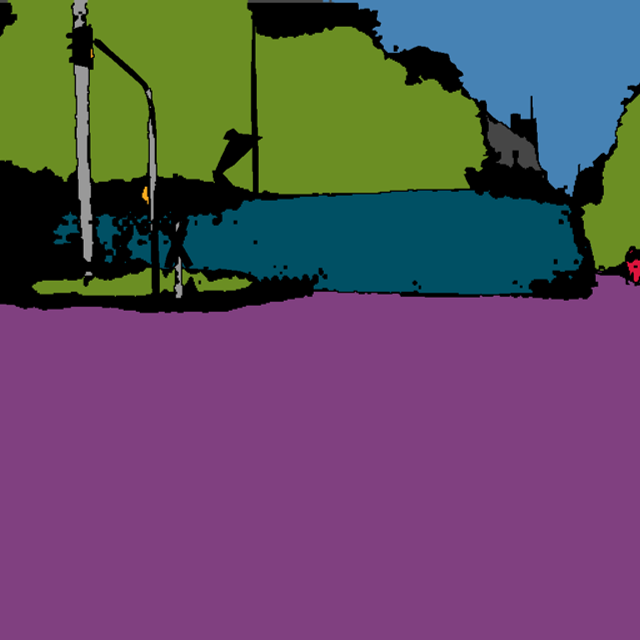}}\\
\rotatebox[origin=c]{90}{RGB}&
\raisebox{-0.5\height}{\includegraphics[width=0.115\textwidth]{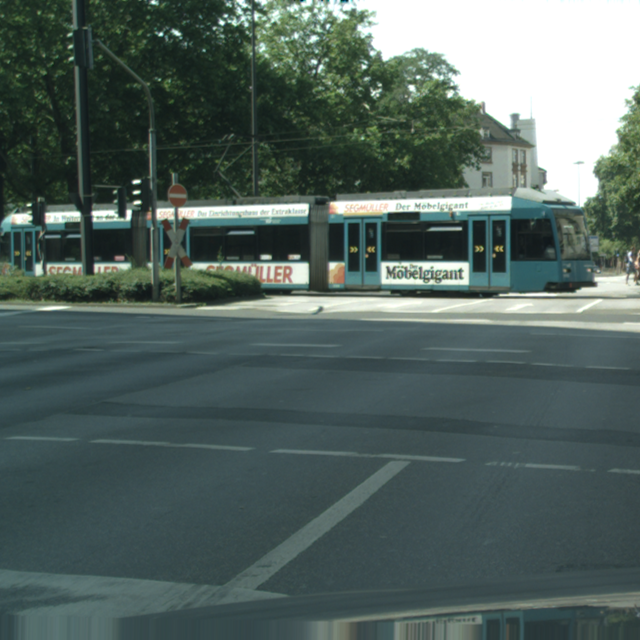}}&
\rotatebox[origin=c]{90}{RGB-\textcolor{gray!40}{D}} &
\raisebox{-0.5\height}{\includegraphics[width=0.115\textwidth]{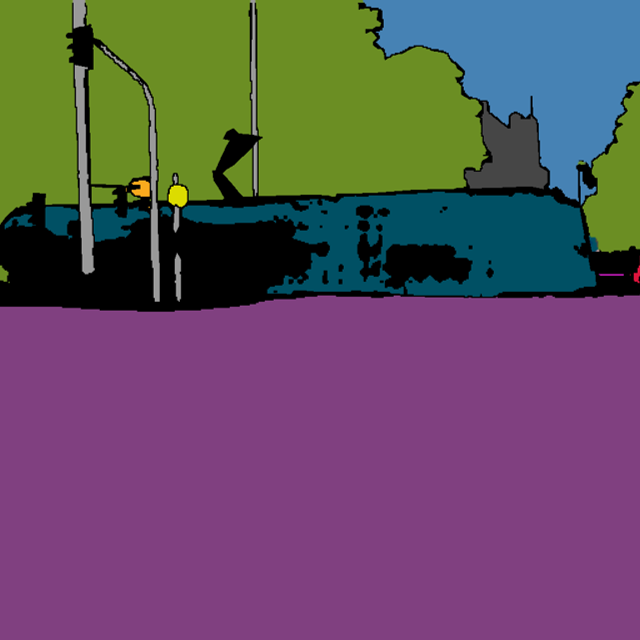}}&
\raisebox{-0.5\height}{\includegraphics[width=0.115\textwidth]{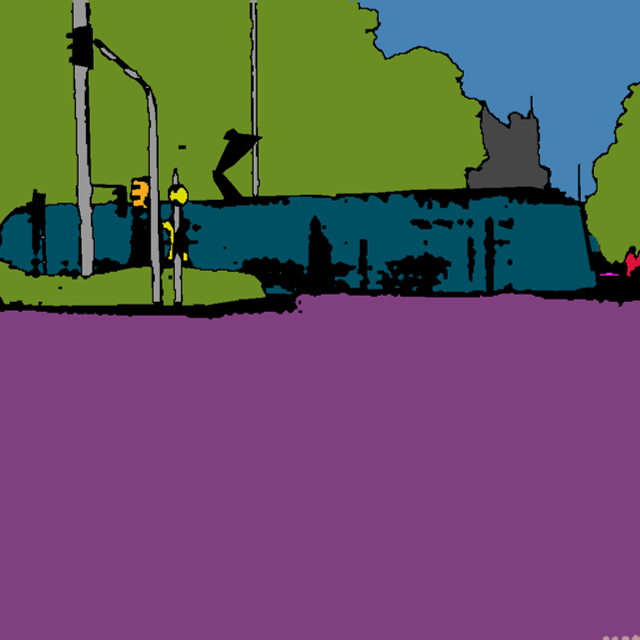}} &
\raisebox{-0.5\height}{\includegraphics[width=0.115\textwidth]{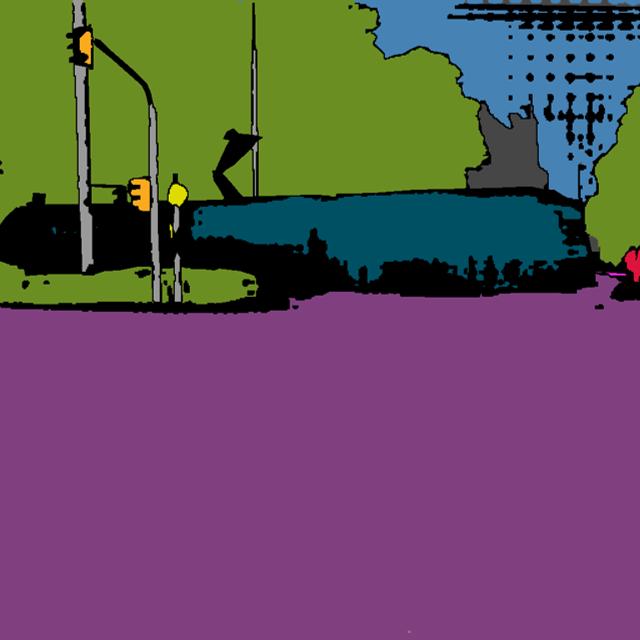}} &
\raisebox{-0.5\height}{\includegraphics[width=0.115\textwidth]{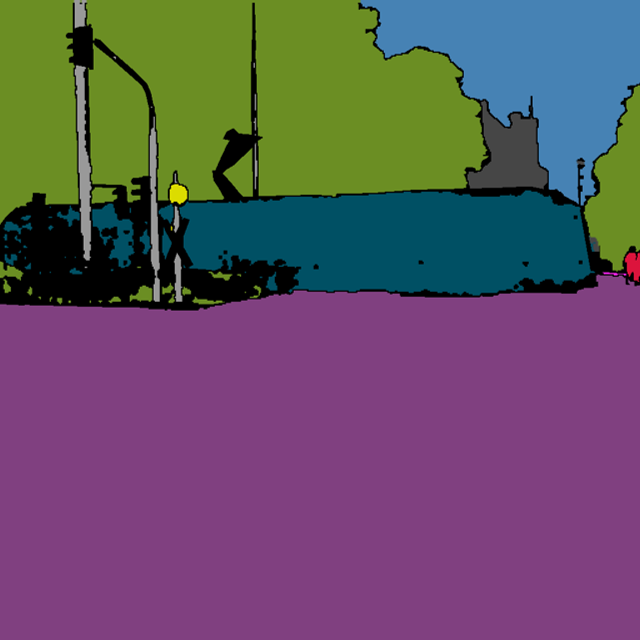}} &
\raisebox{-0.5\height}{\includegraphics[width=0.115\textwidth]{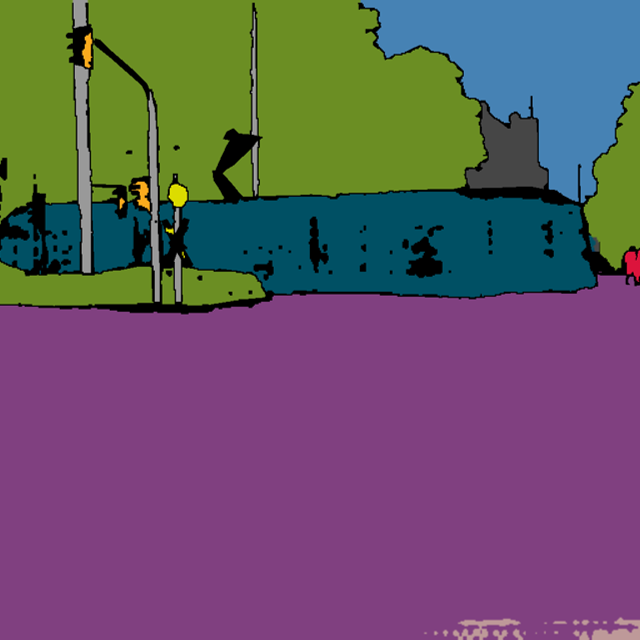}} &
\raisebox{-0.5\height}{\includegraphics[width=0.115\textwidth]{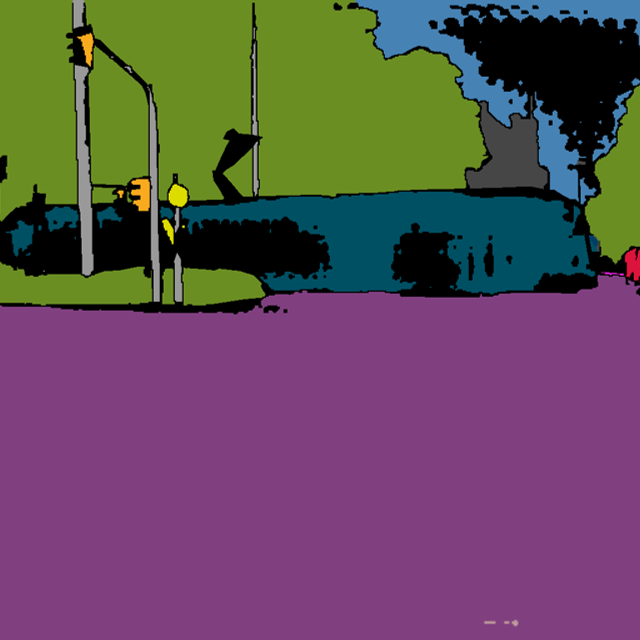}} &
\raisebox{-0.5\height}{\includegraphics[width=0.115\textwidth]{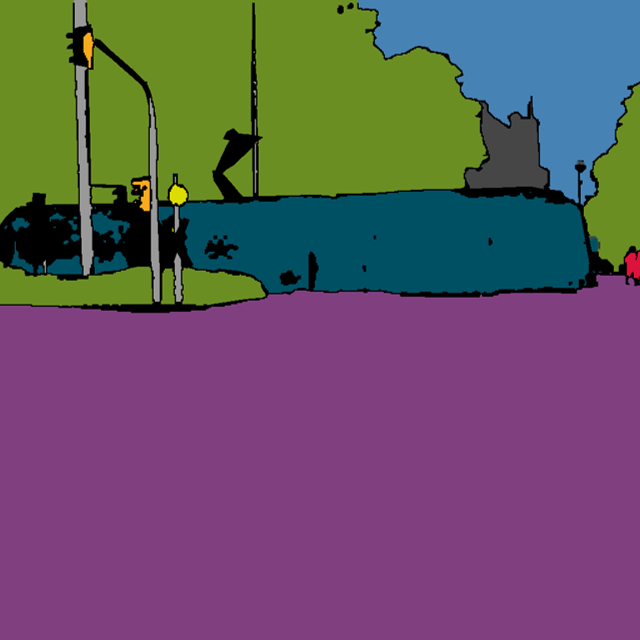}}\\
&Sample Pair&&Fine-tuning& VPT & AdaptFormer & FPT (ours) &VPT$^{\dag}$ & AdaptFormer$^{\dag}$ & FPT$^{\dag}$ (ours)\\
\end{tabular}     
\caption{\textbf{Visualization} of samples with missing modalities. The initial two rows are the semantic segmentation results of a sample under system-level failures (\ie, RGB- or Depth-missing) and sensor-level failures (\ie, \textit{rain} and \textit{over-exposure} conditions) 
on the DeLiVER benchmark~\cite{zhang2023delivering}. The final two rows are system-level failures from the Cityscapes dataset~\cite{cordts2016cityscapes}. Black regions mean incorrect predictions. %
}
\label{fig:miss_vis}
\vskip -4ex
\end{figure*}

\noindent\textbf{Visualization of missing-aware segmentation.} 
Fig.~\ref{fig:miss_vis} visualizes the semantic segmentation results of two samples under different MISS cases, both affected by system-level failures (\ie, RGB- or Depth-missing). The first two rows show results from DeLiVER under sensor-level failures (\textit{rain} and \textit{over-exposure}), while the last two rows depict normal sensor conditions from Cityscapes. Despite training with complete modalities, recognizing expansive backgrounds, such as the \textit{sky} and \textit{vegetation}, remains challenging. Additionally, other methods fail to identify nearby elements, such as the \textit{sidewalk} and \textit{train}. In contrast, our FPT, trained with MMS, successfully discerns detailed scene elements.

\section{Conclusions}
\input{tex/conclusion}

\noindent\textbf{Acknowledgement.} {This work was supported in part by Helmholtz Association of German Research Centers, in part by the MWK through the Cooperative Graduate School Accessibility through AI-based Assistive Technology (KATE) under Grant BW6-03. This work was partially performed on the HoreKa supercomputer funded by the MWK and by the Federal Ministry of Education and Research.}

\bibliographystyle{IEEEtran}
\bibliography{main}
\end{document}

%% file: tex/abstract.tex
Integrating information from multiple modalities enhances the robustness of scene perception systems in autonomous vehicles, providing a more comprehensive and reliable sensory framework. However, the modality incompleteness in multi-modal segmentation remains under-explored. In this work, we establish a task called \textbf{Modality-Incomplete Scene Segmentation (MISS)}, which encompasses both system-level modality absence and sensor-level modality errors. To avoid the predominant modality reliance in multi-modal fusion, we introduce a \textbf{Missing-aware Modal Switch (MMS)} strategy to proactively manage missing modalities during training. Utilizing bit-level batch-wise sampling enhances the model's performance in both complete and incomplete testing scenarios. Furthermore, we introduce the \textbf{Fourier Prompt Tuning (FPT)} method to incorporate representative spectral information into a limited number of learnable prompts that maintain robustness against all MISS scenarios. Akin to fine-tuning effects but with fewer tunable parameters ($1.1\%$). Extensive experiments prove the efficacy of our proposed approach, showcasing an improvement of $5.84\%$ mIoU over the prior state-of-the-art parameter-efficient methods in modality missing. The source code is publicly available at \url{https://github.com/RuipingL/MISS}.

%% file: tex/introduction.tex
\label{sec:intro}

\begin{figure}[!t]
	\centering
    \begin{minipage}{.99\linewidth}
        \begin{subfigure}[t]{.99\linewidth}
            \includegraphics[width=1.0\linewidth]
            {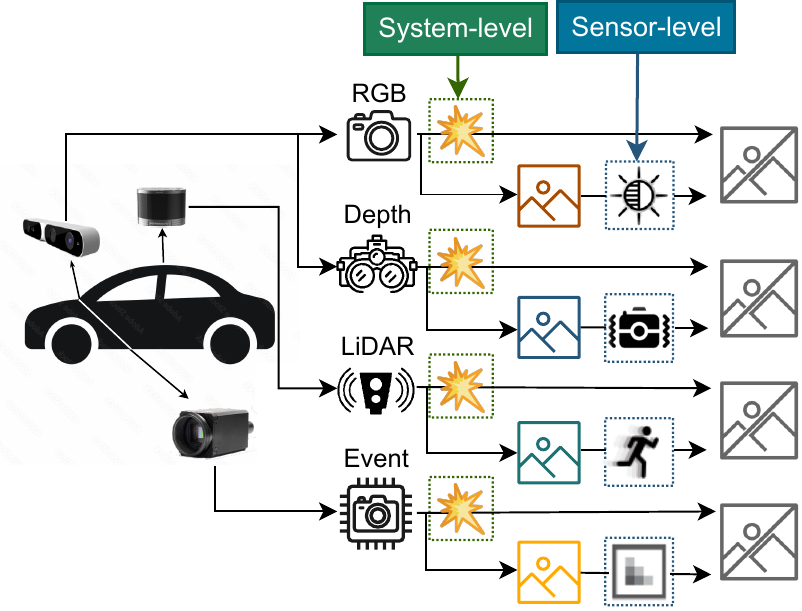}
            \vskip -1ex
            \caption{\small \textbf{Modality-incomplete scenarios} of multi-modal perception in intelligent vehicles, 
            including system-level (\ie, missing modalities) and sensor-level failures (\eg, blurry or misaligned). %
            }
            \label{fig:internal_failure}
        \end{subfigure} \\
        \vskip 1ex
        \begin{subfigure}[b]{.99\linewidth}
            \includegraphics[width=1.0\linewidth]{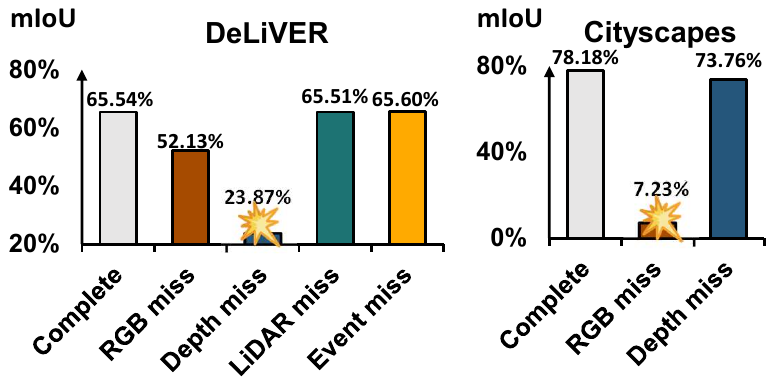}     
            \vskip -1ex
            \caption{\small \textbf{Performance degradation} caused by missing modalities. %
            }
            \label{fig:depth_missing}
        \end{subfigure} 
    \end{minipage}%
	\caption{\small \textbf{Modality-Incomplete Semantic Segmentation (MISS)} aims to cover (a) modality-incomplete scenarios, \eg, in intelligent vehicles. (b) Predominant modality missing leads to severe performance degradation in models trained on complete data. 
 }
    \vskip -3ex
    \label{fig1:missing_sensor}
\end{figure}

Recent advances in image segmentation 
have led to the development of multi-modal perception systems~\cite{zhang2023cmx, zhang2023delivering, bachmann2022multimae, broedermann2022hrfuser} that integrate information from diverse sensors. Nevertheless, the deployment of real-world applications like perception systems in Intelligent Vehicles (IV) is impeded by two challenges: \textit{modality-incomplete issues} and \textit{resource limitations}. Illustrated in Fig.~\ref{fig:internal_failure}, these modality-incomplete issues encompass both \textit{system-level failures}, where the signal breaks down, resulting in the loss of the entire modality, and \textit{sensor-level failures}, exemplified by phenomena such as blurry images or overexposure. Simultaneously, resource constraints intensify the adaptation challenge for cumbersome multi-modal models in downstream tasks that require high generalization.
This paper tackles these two challenges, charting a course toward more robust perception systems.

To achieve this, we propose a new task called Modality-Incomplete Scene Segmentation (MISS), 
to comprehensively explore the aforementioned system- and sensor-level failures. MISS expands upon our previous work, DeLiVER~\cite{zhang2023delivering}, which addressed only sensor-level failures.
Prior studies enabled models to recognize missing modalities through either training on benchmarks with incomplete modalities~\cite{ma2021smil} or predefining a missing ratio for the training set~\cite{lee2023missing_prompt, wang2023multi,ge2023metabev}.
The missing ratio functions as an additional hyperparameter that requires optimization, and an improper setting has the potential to 
aggravate reliance on the predominant modality.
Our observations, illustrated with arbitrary missing modalities in Fig.~\ref{fig:depth_missing}, indicate a significant fragility in the performance of multi-modal networks for semantic segmentation when a predominant dense modality (\textit{e.g.}, RGB or Depth) is missing. For instance, there is a $41.67\%$ mIoU decrease when Depth is missing on the DeLiVER dataset~\cite{zhang2023delivering} and a $70.95\%$ mIoU decrease when RGB is missing on the Cityscapes dataset~\cite{cordts2016cityscapes}. 
Because of the dense prediction, sparse modalities (LiDAR and Event) exert minimal impact on performance. Thus, it is necessary to treat the dense and sparse modalities differently. 
To mitigate the reliance on modality resulting from an inappropriate missing ratio and to differentiate the utilization of $n$ dense and $m$ sparse modalities, we devise the Missing-aware Modal Switch (MMS) training strategy, which is realized by $n{+}m$ bits. 
When tested on datasets containing missing modalities, our MMS consistently outperforms the strategy~\cite{lee2023missing_prompt}, which has a considerably higher missing ratio, achieving a maximum mIoU improvement of $20\%$. 
Notably, when validated on original sets, our proposed training approach maintains on-par performance with marginal variance (approximately ${\pm}0.5\%$ mIoU) compared to training on complete datasets, ensuring reliable multi-modal perception in real-world systems.

\begin{figure}
    \centering
    \includegraphics[width=1.0\columnwidth]{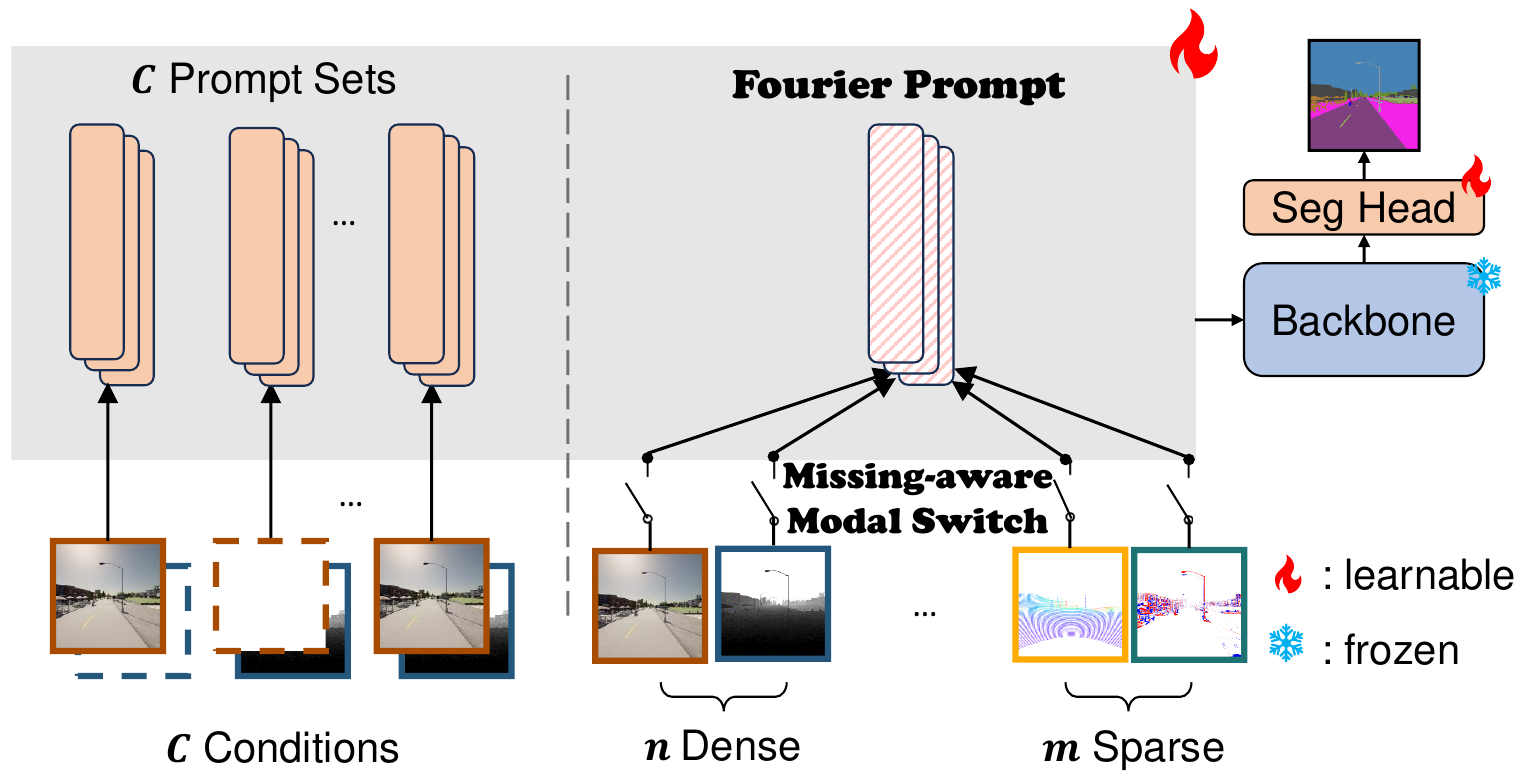}
    \begin{minipage}[t]{.4\columnwidth}
        \vskip -3ex
        \subcaption{Individual prompts~\cite{lee2023missing_prompt}
        }\label{fig2:a}
    \end{minipage}%
    \begin{minipage}[t]{.6\columnwidth}
        \vskip -3ex
        \subcaption{Fourier Prompt}\label{fig2:b}
    \end{minipage}%
    \vskip -1ex
    \caption{\textbf{Paradigms of Prompt Tuning} in semantic segmentation with missing modalities. 
    }
    \vskip -3ex
    \label{fig:paradigms}
\end{figure}

In order to adapt the pre-trained multi-modal network to the downstream tasks while retaining 
general information gathered from pre-training, we adopt prompt tuning~\cite{jia2022vpt}, a parameter-efficient tuning approach.  This approach entails maintaining the frozen state of the pre-trained backbone, with learnable tokens appended to input or feature tokens.
In the prior missing-aware prompt tuning~\cite{lee2023missing_prompt}, as shown in Fig.~\ref{fig2:a}, sets of prompts are assigned to individual missing conditions, and the prompt count increases quadratically with the number of modalities. In contrast, our objective is to formulate a set of robust prompts capable of withstanding all modality-incomplete scenarios, as shown in Fig.~\ref{fig2:b}.
In general, the count of prompt tokens is notably smaller than that of feature tokens ($200$~\textit{v.s.}~${\sim}5000$ in this work). 
Therefore, determining which information should be encoded in the prompt tokens is essential for maximizing the potential of these few learnable parameters. 
Spatial information, crucial for semantic segmentation and susceptible to disruptions from missing data and multiple modalities, becomes impractical to embed within the limited prompt tokens. Conversely, we propose a novel approach, Fourier Prompt Tuning (FPT), that leverages Fast Fourier Transformation (FFT) to extract \textit{global spectral information}.
Previously, the distinct properties of FFTs, global interaction and spectral component extraction, were utilized independently for token mixing~\cite{lee2021fnet, sevim2022fast} and frequency analysis~\cite{guibas2021adaptive, li2023feature}. 
Our FPT takes advantage of both properties by utilizing prompt tokens to identify common frequency components and rectifying them through interaction with all feature tokens. 
The resulting prompt, incorporating global spectral data, effectively complements the spatial characteristics of feature tokens without unnecessary redundancy. 
We conduct a comprehensive series of experiments to establish the robustness of our FPT model over our baseline~\cite{chen2022adaptformer}.
Trained with MMS, our FPT model demonstrates a 
$3.8\%$ mIoU improvement over the baseline in scenarios with complete modalities. Additionally, it obtains a 
gain of $5.84\%$ mIoU in situations involving the absence of the predominant modality. Moreover, our FPT surpasses baselines in all sensor failure cases, achieving a ${\sim}2\%$ mIoU gain compared to the strongest baseline across five failures.

%% file: tex/contributions.tex
To summarize, we present the following contributions:
\begin{compactitem}
    \item A comprehensive task, \textbf{Modality-Incomplete Scene Segmentation (MISS)}, is studied to cover both system-level modality missing and sensor-level modality errors in multi-modal semantic segmentation.  %
    \item We introduce the \textbf{Missing-aware Modal Switch (MMS)} method, using a few bits for modality dropout control in training. It leads to over $20\%$ mIoU enhancement in scenarios lacking the predominant modality while maintaining performance with full modalities.
    \item We propose \textbf{Fourier Prompt Tuning (FPT)}, the first method to inject spectral information into soft prompts via our frequency-spatial cross-attention mechanism, enabling efficient fine-tuning for MISS.
\end{compactitem}

%% file: tex/related.tex
\subsection{Multi-Modal Semantic Segmentation}
Multi-modal semantic segmentation has made significant progress by fusing information from various sensor sources, and current methods in this area are categorized based on their use of different modality combinations.
Some research works
make use of RGB-Depth data for segmentation, \eg, 
ACNet~\cite{hu2019acnet} and SA-Gate~\cite{chen2020sagate}.
RGB-Thermal data
is another established selection 
due to additional information provided by thermal sensors,
\eg, GMNet~\cite{zhou2021gmnet} and ABMDRNet~\cite{zhang2021abmdrnet}.
Some other works~\cite{xiang2021polarization,yan2021nlfnet}
explore RGB-Polarization fusion to enhance material discrimination.
Event cameras are leveraged in~\cite{zhang2021issafe,zhang2021exploring}
to provide high temporal resolution.
Omnivore~\cite{girdhar2022omnivore} and OmniVec~\cite{srivastava2023omnivec} explore fusing images and various data, whereas CMX~\cite{zhang2023cmx} unifies cross-modal RGB-X fusion. 
Recent DeLiVER~\cite{zhang2023delivering} and SegMiF~\cite{liu2023multi} tackle multi-modal fusion in adverse scenarios.  
In this work, we look into the robustness of arbitrary-modal segmentation models under modality-incomplete scenarios, which is more challenging compared to one with complete modalities. 

\subsection{Missing Modality}
In practical situations, sensor malfunctions can lead to the absence of sensor data, which poses a significant challenge for multi-modal semantic segmentation. This challenge starts to grasp attention from the community~\cite{wang2023multi,maheshwari2023missing,wang2023unibev}.
MetaBEV~\cite{ge2023metabev} proposes a BEV-Envolving encoder and switch modality training to alleviate the negative effect brought by the sensor failure for 3D detection and map segmentation.
A multi-modal teacher with a masked modality learning method is proposed by~\cite{maheshwari2023missing} to address missing modalities for semi-supervised segmentation. 
Knowledge distillation is employed by Wang~\etal~\cite{wang2023learnable_distillation_missing} to alleviate the effect brought by missing modalities.
Reza~\etal~\cite{reza2023robust} propose low-rank adaptation and modulation of intermediate features to address missing modalities for RGB-Thermal and -Depth segmentation. 
Chen~\etal~\cite{chen2023redundancy} adopt redundancy-adaptive multi-modal learning to reduce information redundancy considering different modalities. 
A multi-modality guidance network~\cite{zhao2023multimodality_guidance} is proposed to handle missing modalities. 
MedPrompt~\cite{chen2023medprompt} leverages modality translation to alleviate the influence brought by the missing modalities. 
In this work, we tackle the scenarios with incomplete modalities through missing-aware modal switching and prompt tuning.

\subsection{Parameter-Efficient Learning}
Parameter-efficient learning refers to the optimization process where model parameters are systematically adjusted with minimal computational resources. 
Existing works can be grouped into several directions, \eg, parameter-efficient architecture and parameter-efficient tuning~\cite{jia2022vpt, ViPT, yoo2023gvpt}.
Considering parameter-efficient architecture, knowledge distillation~\cite{liu2022transkd, xu2023constructing},
quantization~\cite{choukroun2019low, rokh2023comprehensive},
and the calculation with less parameters, \eg, Fourier Transformation~\cite{lee2021fnet, aich2023efficient, jia2023fourier},
are commonly used. 
Lee~\etal~\cite{lee2021fnet} employ Fourier Transformation in convolution to reduce model parameters.

Considering parameter-efficient tuning, Chen~\etal~\cite{chen2022adaptformer} propose AdaptFormer to achieve task adaptation where only a few parameters are added into the model. 
SSF~\cite{lian2022scaling} performs learnable linear transformation after each frozen operation for parameter-efficient tuning.
With the rapid development of prompt engineering, parameter-efficient prompt tuning grasps massive attention~\cite{jia2022vpt, ViPT, yoo2023gvpt}.
Wang~\etal~\cite{wang2023multitask} propose multi-task prompt tuning.
Nie~\etal~\cite{nie2023pro} propose Pro-tuning to unify prompt tuning for diverse visual tasks.
ViPT~\cite{ViPT} is designed to achieve a visual prompt for multi-modal tracking. 
In this work, we propose Fourier Prompt Tuning for parameter-efficient arbitrary-modal segmentation with missing modalities, considering both spatial and spectral information.
%

%% file: tex/setup.tex
\subsection{Datasets}

\noindent\textbf{DeLiVER}~\cite{zhang2023delivering} contains four distinct modalities: namely, {RGB}, {Depth}, {Event}, and {LiDAR}.
Five sensor failure cases (\textit{motion blur} (MB), \textit{over-exposure} (OE), \textit{under-exposure} (UE), \textit{LiDAR-jitter} (LJ), and \textit{event low-resolution} (EL)) and five weathers (\textit{cloudy}, \textit{rainy}, \textit{sunny}, \textit{foggy}, and \textit{night}) are considered for adverse conditions. 
It has $3983/2005/1897$ images for training/validation/testing at the resolution of $1042{\times}1042$ with $25$ semantic classes. 

\noindent\textbf{Cityscapes}~\cite{cordts2016cityscapes} 
comprises $5000$ images of normal urban scenes, annotated with $19$ classes, $2975/500/1525$ for training/validation/testing. Each image has a size of $1024{\times}2048$. %
We calculate the depth maps with the given sets of disparities and camera parameters. 

\subsection{Implementation Details}
In the context of comparing our Fourier Prompt Tuning with other parameter-efficient training approaches, we utilize the basic version of Vision Transformer (ViT)~\cite{dosovitskiy2021vit} pre-trained with MultiMAE~\cite{bachmann2022multimae} in combination with a ConvNeXt decoder~\cite{liu2022convnext}. 
We further assess the efficacy of the MMS training strategy on the multi-stream multi-modal network, CMNeXt~\cite{zhang2023delivering}. The batch size per GPU is set to be $1$ for MultiMAE and $2$ for CMNeXt. 
Given that MMSs are governed by stochastic bit sequences, we establish a fixed seed to mitigate potential stochasticity-induced impacts. 
The optimizer for all experiments is AdamW. 
Comprehensive configurations are provided in the supplementary.
The images are resized to $768{\times}768$ for DeLiVER and $512{\times}1024$ for Cityscapes. 
Regarding the prompt tuning counterparts, they adhere to their optimal configurations, wherein the count of prompt tokens is $200$ to accommodate diverse scenarios within the realm of MISS.

%% file: tables/deliver.tex
\begin{table}
\footnotesize
\centering
\caption{\small \textbf{Results on the DeLiVER dataset} with \textcolor{gray!40}{missing modalities}. $\dag$ denotes our MMS method, while $\ddag$ follows~\cite{lee2023missing_prompt}. 
}    
\vskip -1ex
\label{tab:deliver}
\resizebox{\columnwidth}{!}{
\setlength{\tabcolsep}{1mm}{

\begin{tabular}{l|l|l|l|l}
\toprule
Method &\#Params (M)& RGB-Depth & \textcolor{gray!40}{RGB}-Depth & RGB-\textcolor{gray!40}{Depth}\\
\midrule\midrule
Full Fine-tuning &96.16& 58.94 & 38.55 & 24.60\\
\midrule
Decoder tuning&09.92& 50.74 &22.04 &25.72\\ 
+ Gated VPT~\cite{yoo2023gvpt}&+0.16&55.71&23.66&25.53\\
+ VPT Deep~\cite{jia2022vpt}&+1.85&56.00&24.04&26.22\\
+ Missing-P~\cite{lee2023missing_prompt}&+3.04&56.08&25.51&24.23\\
+ AdaptFormer~\cite{chen2022adaptformer}&+1.19&55.84&26.54&25.52\\
\rowcolor[gray]{.9} + FPT (ours)&+1.07 \gbf{-0.12}&\bf57.81 \gbf{+1.97}&\bf29.36  \gbf{+2.82}&\bf26.61 \gbf{+1.09}\\
\midrule
+ Gated VPT$^{\dag}$~\cite{yoo2023gvpt}&+0.16&53.66&37.44&48.15\\
+ VPT-Deep$^{\dag}$~\cite{jia2022vpt}&+1.85&54.27&38.46&49.04\\
+ Missing-P$^{\ddag}$~\cite{lee2023missing_prompt}&+3.04&55.31&39.52&04.34\\
+AdaptFormer$^{\dag}$~\cite{chen2022adaptformer}&+1.19&55.74&39.33&47.89\\
\rowcolor[gray]{.9}+ FPT (ours)$^{\dag}$ &+1.07 \gbf{-0.12} &\bf57.38 \gbf{+1.64}&\bf39.60 \gbf{+0.27}&\bf50.73 \gbf{+2.84}\\


\bottomrule 
\end{tabular}
}
}
\end{table}


%% file: tables/cityscapes.tex
\begin{table}
\footnotesize
\centering
\caption{\small \textbf{Results on the Cityscapes dataset} with \textcolor{gray!40}{missing modalities}. $\dag$ denotes our MMS method.}    
\vskip -1ex
\label{tab:cityscapes}
\resizebox{\columnwidth}{!}{
\setlength{\tabcolsep}{1mm}{
\begin{tabular}{l|l|l|l|l}
\toprule
Method &\#Params (M)& RGB-Depth & \textcolor{gray!40}{RGB}-Depth & RGB-\textcolor{gray!40}{Depth}\\
\midrule\midrule
Full Fine-tuning &96.16 &78.18&07.23&64.91\\
\midrule
Decoder tuning&09.92&60.63&04.14&54.62\\ 
+ VPT Deep~\cite{jia2022vpt}&+1.85&71.32&05.36&64.12\\
+ AdaptFormer~\cite{chen2022adaptformer}&+1.19&71.60&\bf05.42&\bf64.37\\
\rowcolor[gray]{.9}+ FPT (ours)&+1.07 \gbf{-0.12}&\bf75.16 \gbf{+3.56}&04.98 \rbf{(-0.44)}&64.22 \rbf{(-0.15)}\\
\midrule
+ VPT-Deep~\cite{jia2022vpt}$^{\dag}$&+1.85&71.25&32.04&69.66\\
+ AdaptFormer~\cite{chen2022adaptformer}$^{\dag}$&+1.19&71.67&33.68&70.47\\
\rowcolor[gray]{.9}+ FPT (ours)$^{\dag}$ &+1.07 \gbf{-0.12}&\bf75.47 \gbf{+3.80}&\bf39.52 \gbf{+5.84}&\bf73.25 \gbf{+2.78}\\ 
\bottomrule 
\end{tabular}
}
}
\vskip -4ex
\end{table}

%% file: tables/two_modalities.tex
\begin{table}
\footnotesize
\centering
\caption{\small \textbf{Analysis of Modality Missing Switch (MMS)} with various architectures, different \textcolor{gray!40}{missing modalities}, and two datasets. $\dag$ denotes our MMS method, while $\ddag$ follows~\cite{lee2023missing_prompt}. 
}    
\vskip -1ex
\label{tab:2_modal}
\resizebox{\columnwidth}{!}{
\renewcommand{\arraystretch}{1}
\setlength{\tabcolsep}{1mm}{
\begin{tabular}{l|c|c|c|c|c|c}
\toprule
\multicolumn{1}{c|}{\multirow{2}{*}{\textbf{Method}}}&\multicolumn{3}{c|}{\textbf{DeLiVER}}&\multicolumn{3}{|c}{\textbf{Cityscapes}}\\ 
& {RGB}-Depth & \textcolor{gray!40}{RGB}-Depth & RGB-\textcolor{gray!40}{Depth}& RGB-Depth & \textcolor{gray!40}{RGB}-Depth & RGB-\textcolor{gray!40}{Depth}\\
\midrule\midrule
MultiMAE&\bf{58.94}&38.55&24.60&\bf78.18&07.23&64.91\\ 
MultiMAE$^{\ddag}$&56.62&42.54&19.24&75.64&40.22&73.54\\
MultiMAE$^{\dag}$&58.68&\bf{45.62}&\bf{52.46}&77.92&\bf49.25&\bf76.37\\
\midrule
CMNeXt&61.93&48.53&32.97&\bf78.69&07.57&73.76\\
CMNeXt$^{\ddag}$&61.13&47.82&33.04&77.09&51.03&75.98\\
CMNeXt$^{\dag}$&\bf{62.24}&\bf{53.73}&\bf{53.39}&78.29&\bf55.07&\bf76.99\\

\bottomrule 
\end{tabular}
}
}
\vskip -2ex
\end{table}

%% file: tables/four_modalities.tex
\begin{table}
\footnotesize
\centering
\caption{
\small \textbf{Results of quad-modal segmentation models}, including training with complete modalities, and training with missing dense (d) and/or sparse (s) modalities.
}    
\vskip -1ex
\label{tab:4_modal}
\resizebox{\columnwidth}{!}{
\renewcommand{\arraystretch}{1}
\setlength{\tabcolsep}{1mm}{
\begin{tabular}{l|c|c|c|c|c}
\toprule
Method & Complete & RGB miss & Depth miss & LiDAR miss& Event miss\\
\midrule\midrule
CMNeXt& \textbf{65.54}& 52.13&23.87&\textbf{65.51}&\textbf{65.60}\\ \midrule
+ Our MMS (d)& 64.81&57.35&55.64&64.75&64.78\\
+ Our MMS (d+s)&65.09&\textbf{57.79}&\textbf{55.80}&65.09&65.14\\ 
\textit{w.r.t. CMNeXt} &~\rbf{-0.45}&~\gbf{+5.66}&~\gbf{+31.93}&~\rbf{-0.42}&~\rbf{-0.46}\\
\bottomrule 
\end{tabular}
}
}
\vskip -2ex
\end{table}

%% file: tables/Ablation.tex
\begin{table}[]
    \footnotesize
    \centering
    \caption{\small \textbf{Ablation study of FPT} based on prompt space.}
    \vskip -1ex
    \resizebox{\columnwidth}{!}{
    \renewcommand{\arraystretch}{.9}
    \begin{tabular}{c c|c|c|c}
        \toprule
         \multicolumn{2}{c|}{Prompt space}& \multirow{2}{*}{RGB-Depth} & \multirow{2}{*}{\textcolor{gray!40}{RGB}-Depth} & \multirow{2}{*}{RGB-\textcolor{gray!40}{Depth}}\\
         Spectrum & Spatiality &&&\\
        \midrule\midrule
        \checkmark&&56.19&38.58&49.95\\
        &\checkmark&56.09&39.19&50.27\\
        \checkmark&\checkmark&\bf57.38&\bf39.60&\bf50.73\\
    \bottomrule

    \end{tabular}
    }
    \label{tab:ablation_study}
\end{table}

%% file: tables/conditions.tex
\begin{table*}
\footnotesize
\centering
\caption{\textbf{Results of MISS task on \textsc{DeLiVER} dataset}. The sensor-level failures are \textbf{MB}: motion blur; \textbf{OE}: over-exposure; \textbf{UE}: under-exposure; \textbf{LJ}: LiDAR-jitter; and \textbf{EL}: event low-resolution. The system-level failures includes \textcolor{gray!40}{RGB} missing and \textcolor{gray!40}{Depth} missing.}
\vskip -1ex
\label{tab:res_condition}
\resizebox{\textwidth}{!}{
\renewcommand{\arraystretch}{.9}
\setlength{\tabcolsep}{1mm}{
\begin{tabular}{lc|ccccc|ccccc|cc|c} 
\toprule
\textbf{Method} & \textbf{\footnotesize{\#Params(M)}} & \textbf{Cloudy} & \textbf{Foggy} & \textbf{Night} & \textbf{Rainy} & \textbf{Sunny} & \textbf{MB} & \textbf{OE} & \textbf{UE} & \textbf{LJ} & \textbf{EL} &\textcolor{gray!40}{RGB}-Depth&RGB-\textcolor{gray!40}{Depth}& \textbf{Mean}  \\
\midrule
Full Fine-tuning&96.16&62.00&58.59&56.41&58.03&60.68&56.94&59.72&52.77&58.97&59.47&38.55&24.60&58.94\\
\midrule
Decoder tuning&09.92&52.85&50.32&44.68&51.31&54.43&50.59&48.98&34.64&49.71&50.25&22.04&25.72&50.74\\

+VPT-Deep&+1.85&58.86&56.14&50.59&56.42&59.34&54.82&54.58&43.42&53.48&56.73&24.04&26.22&56.00\\
+VPT-Deep$^{\dag}$&+1.85&57.01&54.07&50.95&53.35&57.97&52.02&51.99&45.50&50.47&54.42&38.46&\underline{49.04}&54.27\\
+AdaptFormer&+1.19&58.94&55.19&51.40&\underline{56.64}&58.52&54.02&54.25&44.84&54.20&56.92&26.54&25.52&55.84\\
+AdaptFormer$^{\dag}$&+1.19&59.47&55.15&51.73&55.41&58.65&54.17&53.64&47.37&52.87&55.87&\underline{39.33}&47.89&55.74\\
\rowcolor[gray]{.9}+FPT (ours)&+1.07&\bf60.92&\bf57.67&\underline{54.39}&\bf57.58&\bf60.07&\bf56.70&\bf57.60&\underline{47.62}&\bf57.71&\bf58.49&29.36&26.61&\bf57.81\\
\rowcolor[gray]{.9}+FPT (ours)$^{\dag}$&+1.07&\underline{60.29}&\underline{56.43}&\bf54.51&56.48&\underline{59.55}&\underline{55.30}&\underline{56.96}&\bf50.58&\underline{56.83}&\underline{57.27}&\bf39.60&\bf50.73&\underline{57.38}\\

\bottomrule
\end{tabular}
}
}
\vskip -4ex
\end{table*}

%% file: tex/conclusion.tex
In this paper, we look into Modality-Incomplete Scene Segmentation (MISS) in multi-modal semantic scene understanding systems with both system-level modality absence and sensor-level modality outage.
We approach the challenging MISS by proposing a Missing-aware Modal Switch (MMS) solution to govern the presence or absence of each modality during training for MISS.
A Fourier Prompt Tuning (FPT) module is designed with frequency-spatial cross-attention for parameter-efficient fine-tuning while extracting multi-modal complementary cues against MISS.
Extensive experiments on both DeLiVER and Cityscapes benchmarks demonstrate the efficacy of the proposed methods.